\pdfoutput=1
\documentclass[11pt]{article}

\usepackage{EMNLP2023}
\usepackage{times}
\usepackage{latexsym}
\usepackage{booktabs}
\usepackage{multirow}
\usepackage{subcaption}
\usepackage{graphicx}
\usepackage{xcolor}
\usepackage[T1]{fontenc}
\usepackage[utf8]{inputenc}

\usepackage{microtype}

\usepackage{inconsolata}

\usepackage{xspace}

\newcommand{\dataset}{{\sc{seahorse}}\xspace}
\newcommand{\seahorse}{{\sc{seahorse}}\xspace}

\newcommand{\xnli}{{\sc{xnli}}\xspace}
\newcommand{\nli}{{\sc{nli}}\xspace}

\newcommand{\en}{{\texttt{en}}\xspace}
\newcommand{\de}{{\texttt{de}}\xspace}
\newcommand{\es}{{\texttt{es}}\xspace}
\newcommand{\ru}{{\texttt{ru}}\xspace}
\newcommand{\tr}{{\texttt{tr}}\xspace}
\newcommand{\vi}{{\texttt{vi}}\xspace}
 \usepackage{multirow}
\usepackage[normalem]{ulem}
\useunder{\uline}{\ul}{}

\newcommand{\seahorsemodel}{$\textrm{mt5}_\textrm{\seahorse}$\xspace}
\newcommand{\seahorsemodellarge}{$\textrm{mt5\_L}_\textrm{\seahorse}$\xspace}

\newcommand{\nlimodel}{$\textrm{t5}_\textrm{\nli}$\xspace}
\newcommand{\xnlimodel}{$\textrm{mt5}_\textrm{\xnli}$\xspace}
\newcommand{\mfacemodel}{$\textrm{mt5}_\textrm{\sc{mface}}$\xspace}

\title{\textsc{seahorse}: A Multilingual, Multifaceted Dataset for Summarization Evaluation} %for Model and Metric Development}

\author{Elizabeth Clark$^1$\quad
Shruti Rijhwani$^1$\quad
Sebastian Gehrmann$^2$\quad
Joshua Maynez$^1$\\
\textbf{Roee Aharoni}$^2$\quad
\textbf{Vitaly Nikolaev}$^1$\quad
\textbf{Thibault Sellam}$^1$\quad
\textbf{Aditya Siddhant}$^1$\\
\textbf{Dipanjan Das}$^1$\quad
\textbf{Ankur P. Parikh}$^1$\\
$^1$Google DeepMind\quad\quad
$^2$Google Research\\
Contact: \tt{eaclark@google.com}}

\begin{document}
\maketitle
\begin{abstract}
Reliable automatic evaluation of summarization systems is challenging due to the multifaceted and subjective nature of the task. This is especially the case for languages other than English, where human evaluations are scarce. In this work, we introduce \dataset, a dataset for multilingual, multifaceted summarization evaluation. \dataset consists of 96K summaries with human ratings along 6 dimensions of text quality: comprehensibility, repetition, grammar, attribution, main ideas, and conciseness. \dataset covers 6 languages, 9 systems (including the reference text), and 4 summarization datasets. As a result of its size and scope, \dataset can serve both as a benchmark to evaluate learnt metrics, as well as a large-scale resource for training such metrics.
We show that metrics trained with \dataset achieve strong performance on two out-of-domain meta-evaluation benchmarks: TRUE \cite{honovich-etal-2022-true} and mFACE \cite{mface-2022}. We make the \dataset dataset and metrics publicly available for future research on multilingual and multifaceted summarization evaluation.\footnote{Data and metrics are available at \url{https://goo.gle/seahorse}}
\end{abstract}

\section{Introduction}

Evaluating the quality of generated text is an increasingly difficult problem as large language models produce text of rapidly improving quality~\cite{radford2019language,instruct-gpt-2022,palm-2022}. In spite of the improvements, such models often generate text that includes hallucinations and other subtle errors~\cite{wiseman-etal-2017-challenges,maynez-etal-2020-faithfulness,parikh-etal-2020-totto,ji2023survey,borji2023categorical}, making reliable evaluation essential for driving progress. 

\begin{figure}
\centering
\includegraphics[trim={7.5cm 1.5cm 7.5cm 3.5cm},clip,width=0.48\textwidth]{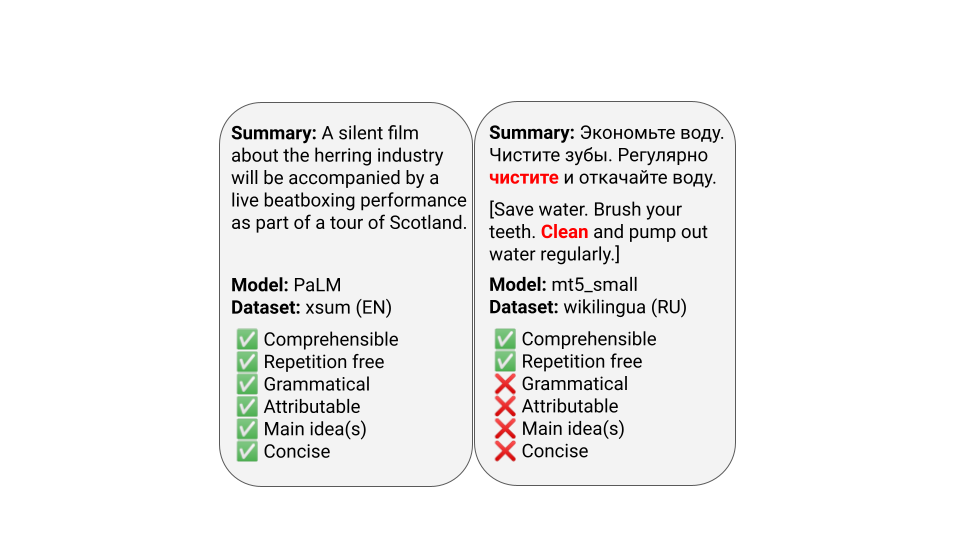}
\caption{Two summaries from the \seahorse dataset paired with human ratings for 6 dimensions of quality. In the second summary, the word in \textbf{\textcolor{red}{bold}} has a grammatical error in Russian; it uses the wrong aspect. The rater has noted this error, along with several others.}
\label{fig:examples}
\end{figure}

Common n-gram metrics such as BLEU~\cite{papineni-etal-2002-bleu} and ROUGE~\cite{lin-2004-rouge} are often not well correlated with human judgments for many natural language generation (NLG) tasks such as machine translation~\cite{kocmi-etal-2021-ship,freitag-etal-2021-experts}, summarization~\cite{kryscinski-etal-2020-evaluating}, and dialogue~\cite{dziri-etal-2022-evaluating}. Consequently, human evaluation is often necessary to reliably evaluate NLG systems. However, designing human annotation pipelines and obtaining annotations is resource-intensive, time-consuming, and not easily reproducible. Developing more reliable automatic evaluation metrics would make model development faster and more efficient. With this in mind, much recent work has focused on learnt metrics, i.e., neural classification or regression models that aim to directly predict scores that evaluate the quality of generated text~\cite{bert-score,sellam-etal-2020-bleurt,rei-etal-2020-comet,gpteval-2023}, often trained with human ratings.

As a result, large-scale collections of human evaluations serve two critical roles in NLG metric development: (1) a source of training data for learnt metrics and (2) a meta-evaluation benchmark for the performance of these learnt metrics. The large potential of such datasets is exemplified by the WMT metrics shared task,\footnote{\url{https://wmt-metrics-task.github.io/}} which has enabled rapid development of learnt metrics for machine translation that exhibit considerably higher correlation to human judgment than BLEU~\cite{bojar-etal-2016-results,freitag-etal-2021-results}.

However, outside of machine translation, the existence of such collections of human judgments is limited. 
Human annotations collected in NLG evaluations are rarely released \citep{repairing-2022}, and even when they are, they tend to cover a single language (typically English) and are from a single dataset or task, limiting the robustness of models and metrics trained on these annotations. Moreover, such annotations are often based on the test split of existing datasets~\cite[e.g.,][]{fabbri-etal-2021-summeval,mface-2022}, which can be problematic for training learnt metrics. This is because the primary advantage of reliable automatic evaluation is to help model development, e.g., hyperparameter selection on the validation set; therefore a neural metric trained on test set annotations would, in general, lead to overfitting.

In this work, we propose \dataset,\footnote{\dataset stands for \textit{SummariEs Annotated with Human Ratings in Six languagEs}.} a large-scale dataset for multilingual summarization evaluation.
Our dataset consists of 96K summaries with ratings along 6 quality dimensions: comprehensibility, repetition, grammar, attribution, main ideas, and conciseness, in 6 languages, for 9 systems (8 models plus the human-authored reference summaries) across 4 summarization datasets (see examples in \autoref{fig:examples}). The training and validation splits of the dataset come from the validation sets of the original summarization corpora to prevent test set contamination when training metrics. This permits us to train a learnt metric for each quality dimension that can be used for offline model evaluation.

We evaluate the metrics learned from \dataset on the \dataset test set, as well as other existing meta-evaluation benchmarks, such as mFACE~\cite{mface-2022} and TRUE~\cite{honovich-etal-2022-true}. Our experiments show that the metrics generalize across datasets, tasks, and languages.
For example, we demonstrate that although \dataset includes data in 6 languages, the resulting learnt metrics achieve strong performance on the mFACE benchmark, which consists of 45 languages, exhibiting their zero-shot multilingual generalization potential. To summarize, the contributions of this paper are:
\begin{itemize}
    \item We conduct a comprehensive, large-scale human evaluation for summarization across six languages, six quality facets, nine systems and four datasets, resulting in over 96K human-rated summaries. To the best of our knowledge, this is the largest multilingual, multifaceted summarization evaluation resource.
    \item We train a learnt metric for each of the evaluated quality facets, and show that the metrics outperform strong baselines across our in-domain test set and previously published out-of-domain benchmarks, highlighting the quality of the human annotations we collect and the broad utility of our learnt metrics.
    \item We release our dataset and metrics to foster future work on multilingual, multifaceted summarization.
\end{itemize}

\section{The \dataset dataset}\label{sec:dataset}
The \dataset dataset consists of 96,645 summaries annotated with human ratings along 6 quality dimensions.
In this section, we describe the \dataset dataset, how we generated the summaries, and how we collected the annotations. 

\begin{table}[tb]
\centering
\begin{tabular}{l l r r}    
\toprule
language & dataset & articles & annotations \\
\midrule
\multirow{2}{*}{de}
 & mlsum & 3359 & 7506 \\
 & wikilingua & 2999 & 7085 \\
\midrule
\multirow{3}{*}{en}
 & xsum & 894 & 6651 \\
 & xlsum & 2433 & 7884 \\
 & wikilingua & 2383 & 7804 \\
\midrule
\multirow{3}{*}{es}
 & xlsum & 2231 & 4890 \\
 & mlsum & 2235 & 4857 \\
 & wikilingua & 2183 & 5002 \\
\midrule
\multirow{2}{*}{ru}
 & xlsum & 3298 & 7254 \\
 & wikilingua & 2948 & 7288 \\
\midrule
\multirow{2}{*}{tr}
 & xlsum & 2186 & 10627 \\
 & wikilingua & 770 & 4791 \\
\midrule
\multirow{2}{*}{vi}
 & xlsum & 2497 & 7522 \\
 & wikilingua & 1951 & 7484 \\
\bottomrule
\end{tabular}
\caption{The number of unique articles and the number of annotated summaries collected from each dataset to create \seahorse.
Each article is summarized by several different summarization systems, which were evaluated by human annotators.
}\label{tab:counts_by_dataset}
\end{table}

\subsection{The summaries}

The examples in \dataset are in 6 languages: German (\de), English (\en), Spanish (\es), Russian (\ru), Turkish (\tr), and Vietnamese (\vi). We chose these languages by considering geographic and typological diversity and the availability of summarization datasets in those languages.

The summaries are based on articles from 4 different datasets in the GEM benchmark \citep{gehrmann-etal-2021-gem}:
\begin{itemize}
    \item \textbf{XSum}~\cite{narayan-etal-2018-dont}: An English dataset where the task is to generate a one-sentence summary of a BBC News article.
    \item \textbf{XL-Sum}~\cite{hasan-etal-2021-xl}: Similar to XSum, the goal of this dataset is to generate a single-sentence summary of a BBC news article, but it covers 44 languages excluding English.
    \item \textbf{MLSum}~\cite{scialom-etal-2020-mlsum}: A summarization dataset obtained from online newspapers in 5 languages.
    \item \textbf{WikiLingua}~\cite{ladhak-etal-2020-wikilingua}: A dataset in 18 languages where the goal is to summarize how-to guides from WikiHow.
\end{itemize}

A breakdown of \dataset across languages and datasets is in \autoref{tab:counts_by_dataset}. 

For each dataset, we randomly selected articles from their validation splits to comprise the \dataset training and validation sets, and articles from the test splits to make up the \dataset test set. This distinction is important when using the dataset for training evaluation metrics (discussed in \S\ref{sec:metrics}), because learnt metrics are typically used for model development, and hyperparameter selection is done on the validation set. Using a metric that was trained on test data would lead to overfitting. Our dataset construction ensures that a learnt metric can be trained on \dataset data without concerns of test set leakage.

Next, we generate summaries for each article in the dataset.
The summaries come from a subset of 9 different systems, which we will denote as follows:

\begin{itemize}
\item \textbf{reference}: The human-authored summaries associated with each article from the original datasets.

\item \textbf{t5\_base}: The 220M-parameter version of the T5 model \citep{t5-2020}. (This model is English-only, so we only use it to generate summaries with our \en datasets.)

\item \textbf{t5\_base\_250}: The t5\_base model with an under-trained checkpoint, trained for only 250 steps (\en only).

\item \textbf{t5\_xxl}: The 11B-parameter version of T5 (\en only).

\item \textbf{mt5\_small}: The 300M-parameter version of mT5 \citep{xue-etal-2021-mt5}.

\item \textbf{mt5\_small\_250}: The same mt5\_small model but using the checkpoint after training 250 steps.

\item \textbf{mt5\_xxl}: The 13B-parameter mT5 model.

\item \textbf{palm\_1shot}: 540B-parameter PaLM model~\citep{palm-2022} prompted with one in-domain example.

\item \textbf{palm\_finetuned}: 540B-parameter PaLM model~\citep{palm-2022} finetuned on training data for the respective dataset.
\end{itemize}

Our choice of systems covers a range of expected system performances in order to capture a large diversity of system outputs and model error types.
For instance, an under-trained small model (\textbf{mt5\_small\_250}) would likely have different errors than a 1-shot large language model (\textbf{palm\_1shot}).
Details about how the summaries are generated from these models are in Appendix \ref{app:training}.

\subsection{Annotation methodology}
\begin{figure*}
\centering
\includegraphics[width=1.0\textwidth,trim={0.25cm 2.5cm 0.25cm 0.5cm},clip]{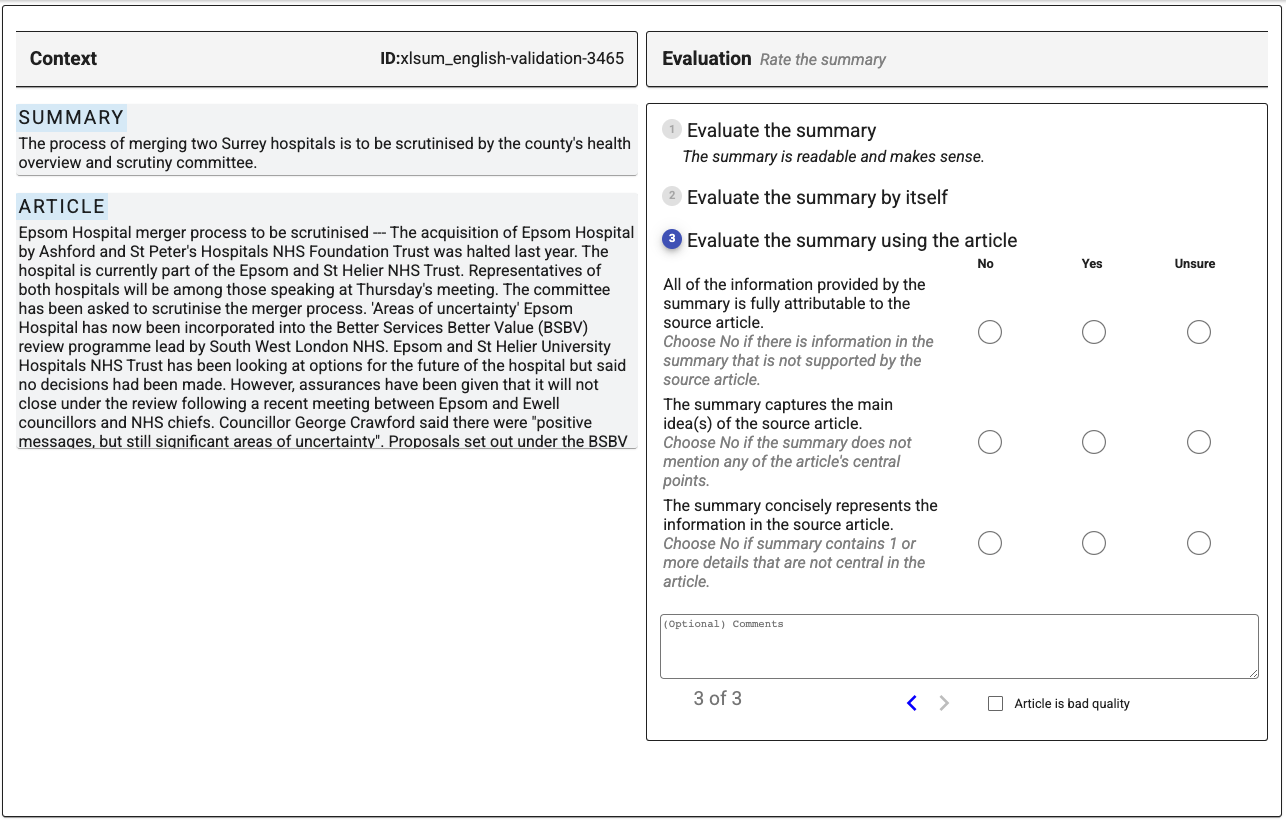}
\caption{The annotation interface used to collect \dataset. First, Question 1 and the summary are shown to the evaluator. Once they confirm that the summary is comprehensible, Questions 2--3 are shown. Finally, the article and Questions 4--6 are displayed (as pictured above).}
\label{fig:cc_interface}
\end{figure*}

For each summary, we collect annotations along 6 dimensions, also referred to as Q1--6:
\begin{itemize}
    \item [\textbf{Q1}] \textbf{comprehensible}: The summary can be read and understood by the rater. (If ``No,'' the rest of the questions will be skipped.)
    \item [\textbf{Q2}] \textbf{repetition}: The summary is free of unnecessarily repeated information.
    \item [\textbf{Q3}] \textbf{grammar}: The summary is grammatically correct.
    \item [\textbf{Q4}] \textbf{attribution}: All the information in the summary is fully attributable to the source article, as defined in \citet{rashkin2021measuring}.
    \item [\textbf{Q5}] \textbf{main ideas}: The summary captures the main idea(s) of the source article.
    \item [\textbf{Q6}] \textbf{conciseness}: The summary concisely represents the information in the source article.
\end{itemize}
For the first 3 questions, annotators see only the summary.
The article is revealed when the raters are answering questions 4--6.
They can answer ``Yes,'' ``No,'' or ``Unsure'' to each question and have the option to leave comments or flag any issues they see in the article.
The annotation interface is shown in \autoref{fig:cc_interface}.

Note that our annotation process is \emph{reference-less}, i.e., the annotator is never comparing a model-generated summary with the reference summary.
They evaluate each summary on its own.
Given the subjectivity of summarization, we believe this approach allows us to adequately reward models that generate relevant summaries that may be different than the reference. 
Moreover, this enables us to train reference-less metrics in \S\ref{sec:metrics}, which have an added benefit of being able to be used at inference time for re-ranking. 

The raters are paid, full-time annotators who were trained for this specific task and worked under the supervision of a project manager.
For the non-English languages, the raters are bilingual, proficient in both the annotation language and English.
They received a detailed set of instructions in English describing the 6 dimensions of quality and positive and negative examples of each in the target language. We created a set of 109 summaries with gold ratings, which we used to train the raters. Each annotator rated 20--30 summaries from this gold set. If the rater performed well on this subset, they were qualified to move forward with the annotation task. Otherwise, the annotator received feedback and were asked to complete another 10--20 ratings. This training process was repeated as needed.

A small number of approved annotators were removed during the annotation process, due to issues flagged by the annotation team and the authors. The ratings from the removed annotators are not included in the dataset.

\section{Dataset analysis}\label{sec:dataset_analysis}
\begin{table}[]
\centering
\begin{tabular}{llll}    
\toprule
model & length & rouge & 20\% copy\\
\midrule
reference & 227 & 20.26 & 0.00 \\
t5\_base\_250 & 92 & 20.95 & 0.00\\
t5\_base & 101 & 22.02 & 0.02\\
t5\_xxl & 115 & 21.65 & 0.01\\
mt5\_small\_250 & 128 & 21.33 & 0.02\\
mt5\_small& 171 & 21.81 & 0.04\\
mt5\_xxl & 194 & 20.77 & 0.01\\
palm\_1shot & 254 & 27.34 & 0.14\\
palm\_finetuned & 194 & 20.97 & 0.01\\
\bottomrule
\end{tabular}
\caption{The median number of characters (\texttt{length}), ROUGE-L between the summary and article (\texttt{rouge}), and the proportion of summaries where the first 20\% of the summary exactly matches the beginning of the source article (\texttt{20\% copy}) for all the summaries generated by each model.}\label{tab:basic_stats_by_model}
\end{table}
\begin{table}[h]
\centering
\scriptsize
\begin{tabular}[t]{lllllll}
\toprule
Model & Q1 & Q2 & Q3 & Q4 & Q5 & Q6 \\ \hline
reference & 0.97 & 0.97 & 0.91 & 0.54 & 0.68 & 0.43 \\
t5\_base\_250 & 0.97 & 0.79 & 0.91 & 0.41 & 0.42 & 0.25 \\
t5\_base & 0.98 & 0.92 & 0.93 & 0.59 & 0.59 & 0.43 \\
t5\_xxl & 0.99 & 0.97 & 0.95 & 0.65 & 0.67 & 0.51 \\
mt5\_small\_250 & 0.71 & 0.43 & 0.59 & 0.27 & 0.19 & 0.1 \\
mt5\_small & 0.86 & 0.57 & 0.73 & 0.36 & 0.35 & 0.19 \\
mt5\_xxl & 0.96 & 0.94 & 0.88 & 0.55 & 0.65 & 0.43 \\
palm\_1shot & 0.88 & 0.85 & 0.79 & 0.71 & 0.57 & 0.47 \\
palm\_finetuned & 0.98 & 0.98 & 0.9 & 0.69 & 0.71 & 0.56 \\
\bottomrule
\end{tabular}
\caption{The percent of ``Yes'' responses, broken down by model and question.}\label{tab:aggregate_yes_rate}
\end{table}
We first analyze the dataset's composition and the quality of the collected annotations.
Table \ref{tab:basic_stats_by_model} contains the median length of summaries produced by each model, along with two measures of the overlap between the summaries and the source articles.
The 1-shot PaLM model is particularly likely to copy from the article as its output, obtaining the highest ROUGE-L\footnote{All ROUGE scores in this paper are calculated with SentencePiece tokens: \url{https://github.com/google/sentencepiece}} \cite{lin-2004-rouge} scores between the summary and the article.
In 14\% of cases, the beginning of the 1-shot summaries (the first 20\% of the summary) exactly matched the beginning of the reference article.

Table \ref{tab:aggregate_yes_rate} shows the percent of summaries from each summarization system that received a positive (i.e., ``Yes'') rating from annotators.
While there is variation across models and datasets, most summaries are rated positively for questions 1--3 (comprehensibility, repetition, and grammar).
The rate of positive responses drops for questions 4--6 (attribution, main ideas, and conciseness), indicating that these areas remain a challenge for summarization models.
A more detailed break down of the positive response rates is in Appendix \ref{app:yes_rate}.

Note that the reference summaries do not always receive the highest rate of positive responses.
The way in which reference texts are collected may limit their quality along some dimensions.
For example, the text that was collected as a reference summary may not have been intended to be read as a standalone replacement for the article, and therefore may not be fully attributable to the article, as \citet{rashkin2021measuring} point out.

We can use the positive response rates to inspect the quality of the dataset by verifying the presence of 3 patterns we expect to see in the data: 1) higher positive response rates for better summarization models, 2) high correlation between the responses to Q4\&6 and Q5\&6, and 3) annotator agreement across the 6 dimensions.

\paragraph{Order of model quality}
Our first expectation is that summaries generated by better summarization models should receive more positive responses from raters.
We have 3 types of model pairs where we can expect one model to generate better summaries than the other: 1) a larger model should outperform a smaller model (the xxl vs.\ the small model), 2) a fully trained model should outperform an under-trained model (the small vs.\ the small\_250 model), and 3) a finetuned model should outperform a 1-shot prompted model (the finetuned vs.\ 1-shot PaLM models).

We compare how often these model pairs produce low-quality summaries, i.e., summaries that are unintelligible to readers. 
In Table \ref{tab:aggregate_yes_rate}, we see that mt5\_xxl produces fewer incomprehensible (Q1) summaries than mt5\_small, which produces fewer than mt5\_small\_250.
The same holds true for the T5 models, and palm\_finetuned produces fewer incomprehensible summaries than palm\_1shot, reflecting the expected relationship in quality between model pairs.
While these results are averaged over the entire dataset, we see the same result when controlling for the source article and only considering items that have summaries generated by all 9 systems (see Appendix~\ref{app:yes_rate}).

This pattern generally holds across the other dimensions of quality as well.
There is one notable exception: PaLM's perfomance on attribution (Q4).
For 4 languages, palm\_1shot is more often rated as being faithful to the input article than palm\_finetuned, which is likely due to its tendency to copy the article directly.

Generally, however, the \seahorse ratings capture the relative differences in model quality we expect to see when evaluating two models with known differences.

\begin{figure}
\centering
\includegraphics[width=.48\textwidth]{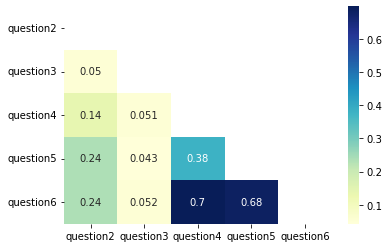}
\caption{The Pearson correlation between responses for questions 2-6.}
\label{fig:q_correlation}
\end{figure}

\paragraph{Correlation between dimensions}
Conciseness (Q6) is related to two other dimensions in our annotation: attribution (Q4) and main ideas (Q5).
A summary cannot be considered a ``concise representation of the information in the article'' if it has information that is not in the article (i.e., a ``No'' response for Q4) or if does not represent the main points in the article (i.e., a ``No'' response for Q5), which was a detail pointed out to evaluators in the task instructions.
Therefore, we expect Q6 to be positively correlated with both of these dimensions if the annotators understood the task and the relationship between the dimensions of quality.

In $>99\%$ of cases when the annotator says a summary is not attributable (Q4) or they say it lacks the main ideas from the article (Q5), they also say it is not concise (Q6).
This is also reflected in \autoref{fig:q_correlation}, which shows that the strongest correlation between questions is between questions 4\&6 and questions 5\&6.
These results show the pattern we expect to see in the data given the task definition and instructions, and it demonstrates the annotators' ability to understand and execute the annotation task.

\begin{table}
\small
\begin{tabular}{llllllll}
\toprule
Lang & Avg & Q1 & Q2 & Q3 & Q4 & Q5 & Q6 \\ 
\midrule
de & 0.84 & 0.97 & 0.98 & 0.95 & 0.81 & 0.67 & 0.66 \\ 
es & 0.82 & 0.92 & 0.97 & 0.83 & 0.74 & 0.7 & 0.74 \\ 
en & 0.81 & 0.97 & 0.94 & 0.95 & 0.69 & 0.61 & 0.69 \\ 
ru & 0.82 & 0.86 & 0.97 & 0.88 & 0.71 & 0.73 & 0.76 \\ 
tr & 0.82 & 0.93 & 0.96 & 0.86 & 0.74 & 0.7 & 0.74 \\ 
vi & 0.81 & 0.95 & 0.98 & 0.88 & 0.68 & 0.66 & 0.69 \\ 
\midrule
avg & 0.82 & 0.93 & 0.97 & 0.89 & 0.73 & 0.68 & 0.72 \\ 
\bottomrule
\end{tabular}
\caption{The average pairwise agreement, broken down by language and question.}
\label{tab:pairwise_agreement_unfiltered}
\end{table}

\begin{table}[]
\centering
\begin{tabular}{llllll}
\toprule
Q1 & Q2 & Q3 & Q4 & Q5 & Q6 \\ 
\midrule
0.49 & 0.87 & 0.35 & 0.47 & 0.4 & 0.41 \\ 
\bottomrule
\end{tabular}
\caption{Krippendorff's $\alpha$ by question.}\label{tab:krippendorff}
\end{table}

\paragraph{Annotator agreement}
While most items in the dataset were annotated once, we collected 2 additional ratings for a subset of the data to compare annotators' scores. 
Out of 8,920 duplicated annotations, the overall pairwise agreement between raters was 82\%. 
\autoref{tab:pairwise_agreement_unfiltered} breaks down the pairwise accuracy across all languages and questions.
Questions 1--3 have higher agreement, while questions 4--6 (which depend on more context and have a higher degree of subjectivity) have lower agreement.
A similar trend is reflected in the Krippendorff's $\alpha$ values \citep[][shown in \autoref{tab:krippendorff}]{Krippendorff1980ContentAA}, which correct for the probability of random agreement, except grammar (Q3) scores lowest.

These patterns in the annotators' responses are positive indicators about the overall quality of the \seahorse ratings.
However, the more important test of the dataset's quality is its usefulness for developing evaluation metrics, which we discuss in the next section.

\section{Learning and evaluating metrics with \seahorse}\label{sec:metrics}

The \dataset dataset is meant to serve both as a source of training data for learnt metrics as well as a meta-evaluation benchmark for these metrics. In this section, we evaluate \dataset on these aspects by looking at how well metrics finetuned with our collected annotations can predict human ratings of generated summaries, both from the \dataset test set and other existing datasets.
When training metrics, we use a filtered version of the dataset that removes all duplicates and non-Yes or No ratings (88,280 total items).
We divide the annotations into train/dev/test splits, where the summaries in the train and dev sets are based on articles from the original datasets' validation sets.
The test set of \seahorse contains summaries of the articles in the original datasets' test sets.

\subsection{Metrics}
One way to train a metric using \dataset is to finetune a text-to-text generation model, where the model is trained to take an article and summary as its input and to output the string `0' or `1' as a prediction of the human rating.
We finetune mT5-\_xxl~\cite{xue-etal-2021-mt5} with the \dataset training set to do this task, finetuning a separate metric for each dimension of quality. We call this model \seahorsemodel\footnote{There are actually 6 different models, one for each question, but we use the notation \seahorsemodel for simplicity.}. More details are in Appendix~\ref{app:training}. Note that our goal is not to train a state-of-the-art metric but rather to evaluate the utility of \seahorse as a resource to train and evaluate such metrics.

We compare the performance of \seahorsemodel to several baselines:
\begin{itemize}
\item \textbf{majority\_class} A majority class baseline (i.e., picking the most frequent class).
\item \textbf{ROUGE-L} The ROUGE-L score between the article and the summary.
\end{itemize}

Specifically for the attribution (Q4) task, we consider a third baseline approach; attribution is closely related to natural language inference (NLI)~\cite{fyodorov2000natural,dagan2006pascal}, and \citet{honovich-etal-2022-true} show that models finetuned on NLI data perform well as faithfulness metrics.
Therefore we consider two variants of an NLI-based baseline:
\begin{itemize}
    \item \textbf{\nlimodel}: An English NLI model proposed by~\citet{honovich-etal-2022-true}.\footnote{\url{https://huggingface.co/google/t5\_xxl\_true\_nli\_mixture}} T5\_xxl is finetuned on the following datasets: SNLI~\cite{bowman-etal-2015-large}, MNLI~\cite{williams-etal-2018-broad}, Fever~\cite{thorne-etal-2018-fever}, Scitail~\cite{khot2018scitail}, PAWS~\cite{zhang-etal-2019-paws}, and VitaminC~\cite{schuster-etal-2021-get}.
    \item \textbf{\xnlimodel}: A multilingual version, where mT5\_xxl is finetuned on XNLI~\cite{conneau-etal-2018-xnli}.
\end{itemize}

We note that since we are operating in the reference-free setting, other learnt metrics such as BLEURT~\cite{sellam-etal-2020-bleurt} or BERTScore~\cite{bert-score} are not applicable since they measure the similarity between the prediction and reference.

We evaluate the \seahorse and baseline metrics in two ways: the area under the ROC curve and the correlation (Pearson's $\rho$) between the metric and human scores. These measures are not sensitive to a thresholding value and are also used in the work we compare with \citep{honovich-etal-2022-true, mface-2022}.

\subsection{Evaluation on the \seahorse test set}
\begin{table*}[]
\centering
\small
\begin{tabular}{l|ll|ll|ll|ll|ll|ll}
\toprule
\multicolumn{1}{c|}{}     & \multicolumn{2}{c|}{Q1}                                       & \multicolumn{2}{c|}{Q2}                                       & \multicolumn{2}{c|}{Q3}                                       & \multicolumn{2}{c|}{Q4}                                       & \multicolumn{2}{c|}{Q5}                                       & \multicolumn{2}{c}{Q6}                                       \\ \hline
Metric                    & \multicolumn{1}{c|}{$\rho$}        & \multicolumn{1}{c|}{roc} & \multicolumn{1}{c|}{$\rho$}        & \multicolumn{1}{c|}{roc} & \multicolumn{1}{c|}{$\rho$}        & \multicolumn{1}{c|}{roc} & \multicolumn{1}{c|}{$\rho$}        & \multicolumn{1}{c|}{roc} & \multicolumn{1}{c|}{$\rho$}        & \multicolumn{1}{c|}{roc} & \multicolumn{1}{c|}{$\rho$}        & \multicolumn{1}{c}{roc} \\ \hline
majority\_class           & \multicolumn{1}{l|}{-}           & 0.5                      & \multicolumn{1}{l|}{-}           & 0.5                      & \multicolumn{1}{l|}{-}           & 0.5                      & \multicolumn{1}{l|}{-}           & 0.5                      & \multicolumn{1}{l|}{-}           & 0.5                      & \multicolumn{1}{l|}{-}           & 0.5                     \\
ROUGE-L                   & \multicolumn{1}{l|}{0.04}          & 0.54                     & \multicolumn{1}{l|}{0.06}          & 0.54                     & \multicolumn{1}{l|}{-0.03}         & 0.43                     & \multicolumn{1}{l|}{0.13}          & 0.55                     & \multicolumn{1}{l|}{0.03}          & 0.53                     & \multicolumn{1}{l|}{0.02}          & 0.54                    \\
\xnlimodel            & \multicolumn{1}{l|}{-}             & -                        & \multicolumn{1}{l|}{-}             & -                        & \multicolumn{1}{l|}{-}             & -                        & \multicolumn{1}{l|}{0.43}          & 0.78                     & \multicolumn{1}{l|}{-}             & -                        & \multicolumn{1}{l|}{-}             & -                       \\
\seahorsemodellarge & \multicolumn{1}{l|}{0.44} & 0.88            & \multicolumn{1}{l|}{0.74} & 0.97            & \multicolumn{1}{l|}{0.37} & 0.81           & \multicolumn{1}{l|}{0.55} & 0.82            & \multicolumn{1}{l|}{0.46} & 0.78            & \multicolumn{1}{l|}{0.45} & 0.77  \\
\seahorsemodel & \multicolumn{1}{l|}{\textbf{0.52}} & \textbf{0.90}            & \multicolumn{1}{l|}{\textbf{0.86}} & \textbf{0.98}            & \multicolumn{1}{l|}{\textbf{0.45}} & \textbf{0.84}            & \multicolumn{1}{l|}{\textbf{0.59}} & \textbf{0.85}            & \multicolumn{1}{l|}{\textbf{0.50}} & \textbf{0.80}            & \multicolumn{1}{l|}{\textbf{0.52}} & \textbf{0.81}           \\ \hline
\end{tabular}
\caption{Metrics' ability to predict \dataset ratings, measured with Pearson's coefficient ($\rho$) and the area under the ROC curve (roc).
\seahorsemodellarge is a finetuned version of mT5\_large; the other mt5 metrics finetune mT5\_xxl.}\label{tab:metric_results_seahorse}
\end{table*}
We first evaluate \seahorsemodel on the \dataset test set to confirm that a model is able to learn to predict the different dimensions of quality in \dataset.
The results are shown in \autoref{tab:metric_results_seahorse}.
As expected, we see that the \seahorsemodel model is able to predict \seahorse ratings better than the baselines according to both our metrics.
The repetition (Q2) metric performs the best out of the 6 dimensions, which is also the dimension with the highest pairwise annotator agreement.
Examples of summaries paired with human, \seahorse, and ROUGE-L ratings can be found in Appendix~\ref{app:seahorse_examples}.

Reducing the size of the base mT5 model from XXL (13B parameters) to Large (1.2B) drops the performance of the metric, but shows similar trends and still outperforms all baseline approaches.
More \seahorsemodellarge results can be found in Appendix~\ref{app:mt5_large}.

\subsection{Evaluation on the mFACE dataset}
\begin{table*}[]
\centering
\scriptsize
\begin{tabular}{l|l|llllll|llllll}
\toprule
\multirow{3}{*}{}                                                                         & \multirow{2}{*}{}         & \multicolumn{6}{c|}{mFACE - 5 languages}                                                                                                                                                                            & \multicolumn{6}{c}{mFACE - all languages}                                                                                                                                                                            \\ \cline{3-14} 
                                                                                          &                           & \multicolumn{2}{c|}{Quality}                                             & \multicolumn{2}{c|}{Attribution}                                         & \multicolumn{2}{c|}{Informativeness}                          & \multicolumn{2}{c|}{Quality}                                             & \multicolumn{2}{c|}{Attribution}                                          & \multicolumn{2}{c}{Informativeness}                           \\ \cline{2-14} 
                                                                                          & Metric                    & \multicolumn{1}{c|}{$\rho$}         & \multicolumn{1}{c|}{roc}           & \multicolumn{1}{c|}{$\rho$}         & \multicolumn{1}{c|}{roc}           & \multicolumn{1}{c|}{$\rho$}        & \multicolumn{1}{c|}{roc} & \multicolumn{1}{c|}{$\rho$}         & \multicolumn{1}{c|}{roc}           & \multicolumn{1}{c|}{$\rho$}         & \multicolumn{1}{c|}{roc}            & \multicolumn{1}{c|}{$\rho$}         & \multicolumn{1}{c}{roc} \\ \hline
\multirow{4}{*}{\textit{\begin{tabular}[c]{@{}l@{}}Not trained \\ on mFACE\end{tabular}}} & majority\_class           & \multicolumn{1}{l|}{-}            & \multicolumn{1}{l|}{0.5}           & \multicolumn{1}{l|}{-}            & \multicolumn{1}{l|}{0.5}           & \multicolumn{1}{l|}{-}           & 0.5                      & \multicolumn{1}{l|}{-}            & \multicolumn{1}{l|}{0.5}           & \multicolumn{1}{l|}{-}            & \multicolumn{1}{l|}{0.5}            & \multicolumn{1}{l|}{-}            & 0.5                     \\
                                                                                          & ROUGE-L                   & \multicolumn{1}{l|}{0.02}           & \multicolumn{1}{l|}{0.51}          & \multicolumn{1}{l|}{0.14}           & \multicolumn{1}{l|}{0.58}          & \multicolumn{1}{l|}{0.06}          & 0.56                     & \multicolumn{1}{l|}{0.06}           & \multicolumn{1}{l|}{0.52}          & \multicolumn{1}{l|}{0.09}           & \multicolumn{1}{l|}{0.52}           & \multicolumn{1}{l|}{0.09}           & 0.52                    \\
                                                                                          & \xnlimodel       & \multicolumn{1}{l|}{-}              & \multicolumn{1}{l|}{-}             & \multicolumn{1}{l|}{0.45}           & \multicolumn{1}{l|}{\textbf{0.82}} & \multicolumn{1}{l|}{-}             & -                        & \multicolumn{1}{l|}{-}              & \multicolumn{1}{l|}{-}             & \multicolumn{1}{l|}{0.34}           & \multicolumn{1}{l|}{0.74}           & \multicolumn{1}{l|}{-}              & -                       \\
                                                                                          & \seahorsemodel & \multicolumn{1}{l|}{\textbf{0.09}}  & \multicolumn{1}{l|}{\textbf{0.73}} & \multicolumn{1}{l|}{\textbf{0.50}}  & \multicolumn{1}{l|}{0.79}          & \multicolumn{1}{l|}{\textbf{0.50}} & \textbf{0.81}            & \multicolumn{1}{l|}{\textbf{0.15}}  & \multicolumn{1}{l|}{\textbf{0.70}} & \multicolumn{1}{l|}{\textbf{0.52}}  & \multicolumn{1}{l|}{\textbf{0.81}}  & \multicolumn{1}{l|}{\textbf{0.40}}  & \textbf{0.74}           \\ \hline
\textit{\begin{tabular}[c]{@{}l@{}}Trained on\\  mFACE\end{tabular}}                      & \mfacemodel    & \multicolumn{1}{l|}{\textbf{0.25*}} & \multicolumn{1}{l|}{0.68}          & \multicolumn{1}{l|}{\textbf{0.51*}} & \multicolumn{1}{l|}{0.81}          & \multicolumn{1}{l|}{0.47}          & 0.79                     & \multicolumn{1}{l|}{\textbf{0.35*}} & \multicolumn{1}{l|}{0.61}          & \multicolumn{1}{l|}{\textbf{0.52*}} & \multicolumn{1}{l|}{\textbf{0.82*}} & \multicolumn{1}{l|}{\textbf{0.47*}} & \textbf{0.80*}          \\ \hline
\end{tabular}

\caption{Metrics' ability to predict mFACE ratings, measured with Pearson's coefficient ($\rho$) and the area under the ROC curve (roc).
The asterisk indicates that the associated model was trained on the training portion of the mFACE dataset.
}\label{tab:metric_results_mface}
\end{table*}
In addition to achieving good performance on the \dataset test set, we would like to evaluate how well models trained on \seahorse generalize to other multilingual summarization human evaluation datasets without any further tuning.
This would give evidence that improving on \dataset would lead to better evaluation metrics in general. 

For this purpose, we choose the mFACE dataset\footnote{We obtained the dataset by contacting the authors.}~\citep{mface-2022}. mFACE contains human evaluations of the XL-Sum test set, which consists of 45 languages on 3 dimensions: quality, attribution, and informativeness. While their definition of attribution is the same as ours (i.e., following AIS~\cite{rashkin2021measuring}), their definitions of quality (\textit{Is the summary comprehensible?}) and informativeness (\textit{Is the summary a good summary of the article?}) do not line up exactly with a single one of our questions, a misalignment that we expect to occur in practice given the lack of standardization of summarization human evaluation.

As a result, for each mFACE dimension, we use the \seahorse metric for the question that is most similar; attribution clearly aligns with Q4, and for quality and informativeness, we consider Q1 and Q6 to be the closest fit, respectively.

We evaluate on both the full mFACE dataset (all languages), as well as the 5-language subset that is common to both mFACE and \seahorse (\en, \es, \ru, \tr, \vi).
In addition to our baseline models, we also compare to an ``upper-bound'' mT5\_xxl model that has been directly trained on mFACE data (\mfacemodel).

Results are shown in Table~\ref{tab:metric_results_mface}.
In all but one column, \seahorsemodel outperforms the other methods that were not trained on the mFACE data and also performs well on the languages it was not finetuned on. 
\seahorsemodel even performs comparably to \mfacemodel on the 5 language subset on all dimensions, and the attribution dimension on the all-language set.
\mfacemodel performs better on quality and informativeness on the all-language set, as one would expect, since it has seen supervised data from those languages and dimensions whereas \seahorsemodel is applied in a zero-shot setting.

\subsection{Evaluation on the TRUE Benchmark}

\begin{table*}[]
\centering
\scriptsize
\begin{tabular}{l|l|l|l|l|l|l|l|l|l|l|l}
\toprule
\multicolumn{1}{c|}{}     & \multicolumn{1}{c|}{FRANK} & \multicolumn{1}{c|}{SummEval} & MNBN          & QAGS-C        & \multicolumn{1}{c|}{QAGS-X} & BEGIN         & \multicolumn{1}{c|}{$Q^2$} & \multicolumn{1}{c|}{DialFact} & Fever          & VitaminC       & PAWS           \\ \hline
majority\_class           & 0.5                        & 0.5                           & 0.5           & 0.5           & 0.5                         & 0.5           & 0.5                        & 0.5                           & 0.5            & 0.5            & 0.5            \\
ROUGE-L                   & 0.55                       & 0.57                          & 0.53          & 0.44          & 0.55                        & 0.63          & 0.54                       & 0.49                          & 0.48           & 0.50           & 0.60           \\
mT5$_{\textrm{\dataset}}$ & \textbf{0.94}              & \textbf{0.87}                 & \textbf{0.83} & \textbf{0.91} & \textbf{0.87}               & 0.84          & 0.82                       & 0.87                          & \textbf{0.91}  & \textbf{0.78}  & \textbf{0.82}  \\
T5$_{\textrm{\nli}}$       & 0.90                       & 0.79                          & 0.76          & 0.77          & 0.85                        & \textbf{0.85} & \textbf{0.83}              & \textbf{0.92}                 & \textbf{0.95*} & \textbf{0.98*} & \textbf{0.99*} \\ \hline
\end{tabular}
\caption{Metrics' performance on the TRUE benchmark, measured with area under the ROC curve. \nlimodel is a T5-xxl model trained on a mixture of NLI datasets that includes the FEVER, VitaminC, and PAWS training sets (and thus those numbers are indicated with an asterisk).}\label{tab:metric_results_q4}
\end{table*}
Finally, we focus on the attribution dimension of quality, since issues of faithfulness in generated text are increasingly important~\cite{wiseman-etal-2017-challenges,tian2019sticking,zhou-etal-2021-detecting,dziri-etal-2022-evaluating,ji2023survey}. The TRUE benchmark~\cite{honovich-etal-2022-true} consists of several English datasets across summarization, dialogue, verification, and paraphrasing: FRANK~\cite{pagnoni-etal-2021-understanding}, SummEval~\cite{fabbri-etal-2021-summeval}, MNBM~\cite{maynez-etal-2020-faithfulness},
QAGS~\cite{wang-etal-2020-asking}, BEGIN~\cite{dziri-etal-2022-evaluating}, $Q^2$~\cite{honovich-etal-2021-q2}, DialFact~\cite{gupta-etal-2022-dialfact}, FEVER~\cite{thorne-etal-2018-fever}, VitaminC~\cite{schuster-etal-2021-get}, and PAWS~\cite{zhang-etal-2019-paws}. 

As in the prior section, we apply \seahorsemodel without any further finetuning to these datasets to assess its ability to evaluate attribution to other datasets and tasks beyond summarization.
In addition to comparing to the majority class and ROUGE-L baselines, we also compare with \nlimodel. 

Results are shown in \autoref{tab:metric_results_q4}. \seahorsemodel achieves the best results across the summarization datasets, which is expected as many of these datasets consist of XSum and CNN/DailyMail~\cite{hermann2015teaching}, the first of which is also a source of the \dataset summaries and the second is a different news summarization dataset. Interestingly, despite only being trained on summarization data, \seahorsemodel performs competitively to \nlimodel on the dialogue datasets (BEGIN, $Q^2$, and DialFact), indicating its suitability for evaluating tasks outside of summarization. \nlimodel performs best on the Fever, VitaminC, and PAWS tasks, which is expected given that the \nlimodel model was trained on these datasets.

\section{Related work}

We briefly review other large-scale datasets of human evaluations of summaries that have been released and compare them to \seahorse, but note that most focus on annotating the test data, which would lead to test data contamination when training metrics.

SummEval~\cite{fabbri-etal-2021-summeval} and RealSumm~\cite{bhandari-etal-2020-evaluating} are summarization meta-evaluation benchmarks with 12,800 and 7,742 annotations respectively. These benchmarks focus on a single language and single dataset: the CNN/DailyMail English summarization dataset.
The RoSE benchmark \citep{liu2022revisiting} contains 22K summary-level annotations across 3 summarization datasets, including a subset from the CNN/DailyMail validation set, and \citet{stiennon2020learning} released 65K summary comparisons on the TL;DR dataset \citep{volske-etal-2017-tl}; however, both only consider English summarization tasks.
\citet{rashkin2021measuring} focus on attribution, releasing $\sim$4.5K annotations from English summarization, table-to-text, and dialogue datasets; \citet{gekhman2023trueteacher} also release attribution annotations for 1.4M summaries, but the labels are machine-generated rather than human-annotated.
GENIE~\citep{khashabi-etal-2022-prompt} released 17K human evaluations across 5 tasks that includes one English summarization task (XSum).

The only other multilingual summarization evaluation dataset, to the best of our knowledge, is mFACE~\cite{mface-2022}, which has annotations for 31,500 summaries covering a broader set of languages (45 languages). mFACE focuses on one dataset (XL-Sum) and a smaller set of models than \seahorse.  
In \S\ref{sec:metrics} we use mFACE as a comprehensive out-of-domain evaluation set, and view it as complementary to \seahorse, which aims to provide large-scale and diverse training data for metrics.

\section{Conclusion}
In this work, we present \seahorse, a large-scale multilingual, multifaceted dataset for summarization consisting of 96K human annotations of summaries. Due to its size and scope, \seahorse enables the training and evaluation of learnt metrics across several quality dimensions. Our results show that \seahorse-trained metrics not only achieve strong performance on our own test set but also generalize to other external and out-of-domain benchmarks: mFACE and TRUE.
In the future, we are interested in exploring how \seahorse can be used more directly to improve the quality of summarization models and metrics, and hope this paper and the public release of \dataset enables further research on these topics.

\section*{Limitations}
The summaries in this work are in 6 languages, and the selection of these languages was based on the number of datasets and articles available for each language.
We would like future work to explore the incorporation of low-resource languages, perhaps with the use of crosslingual and fewshot summarization systems.
While the raters we worked with in this project went through several rounds of instructions and training, there is a degree of subjectivity inherent in the 6 text quality evaluation tasks and human ratings are noisy, as each individual rater may interpret and rate qualities slightly differently.
Finally, the mT5-based metrics presented in this work primarily serve as a demonstration of the potential of the \seahorse data for developing summarization metrics; they have not optimized via thorough hyperparameter search, comparing different modeling architectures or approaches, etc.
We hope the dataset and experimental results will provide a starting point for this type of exploration in the future.

\section*{Ethics Statement}
This work relies on the efforts of human evaluators, who were compensated for their work.
The summaries in this work are machine-generated and should not be treated as truth; they may contain misleading or incorrect information.
None of the human ratings capture this dimension of the text, as our quality dimensions focus on the relationship between the summary and the source article, not a broader set of information or perspectives.
For example, if an article contains a factual error, a summary that contains the same error should be rated as ``Yes'' for Q4 (attribution) because it is consistent with the article.
We used summarization models of varying quality in this work, but all are imperfect and their output should be treated with caution.

\section*{Acknowledgements}
We would like to thank Ashwin Kakarla and his team for help with the annotations, as well as Slav Petrov, Hannah Rashkin, and our EMNLP reviewers for their feedback on the paper.

\bibliography{anthology,custom}

\begin{thebibliography}{56}
\expandafter\ifx\csname natexlab\endcsname\relax\def\natexlab#1{#1}\fi

\bibitem[{Aharoni et~al.(2023)Aharoni, Narayan, Maynez, Herzig, Clark, and
  Lapata}]{mface-2022}
Roee Aharoni, Shashi Narayan, Joshua Maynez, Jonathan Herzig, Elizabeth Clark,
  and Mirella Lapata. 2023.
\newblock \href {https://doi.org/10.18653/v1/2023.findings-acl.220}
  {Multilingual summarization with factual consistency evaluation}.
\newblock In \emph{Findings of the Association for Computational Linguistics:
  ACL 2023}, pages 3562--3591, Toronto, Canada. Association for Computational
  Linguistics.

\bibitem[{Bhandari et~al.(2020)Bhandari, Gour, Ashfaq, Liu, and
  Neubig}]{bhandari-etal-2020-evaluating}
Manik Bhandari, Pranav~Narayan Gour, Atabak Ashfaq, Pengfei Liu, and Graham
  Neubig. 2020.
\newblock \href {https://doi.org/10.18653/v1/2020.emnlp-main.751}
  {Re-evaluating evaluation in text summarization}.
\newblock In \emph{Proceedings of the 2020 Conference on Empirical Methods in
  Natural Language Processing (EMNLP)}, pages 9347--9359, Online. Association
  for Computational Linguistics.

\bibitem[{Bojar et~al.(2016)Bojar, Graham, Kamran, and
  Stanojevi{\'c}}]{bojar-etal-2016-results}
Ond{\v{r}}ej Bojar, Yvette Graham, Amir Kamran, and Milo{\v{s}} Stanojevi{\'c}.
  2016.
\newblock \href {https://doi.org/10.18653/v1/W16-2302} {Results of the {WMT}16
  metrics shared task}.
\newblock In \emph{Proceedings of the First Conference on Machine Translation:
  Volume 2, Shared Task Papers}, pages 199--231, Berlin, Germany. Association
  for Computational Linguistics.

\bibitem[{Borji(2023)}]{borji2023categorical}
Ali Borji. 2023.
\newblock \href {https://arxiv.org/abs/2302.03494} {A categorical archive of
  chat{GPT} failures}.
\newblock \emph{arXiv preprint arXiv:2302.03494}.

\bibitem[{Bowman et~al.(2015)Bowman, Angeli, Potts, and
  Manning}]{bowman-etal-2015-large}
Samuel~R. Bowman, Gabor Angeli, Christopher Potts, and Christopher~D. Manning.
  2015.
\newblock \href {https://doi.org/10.18653/v1/D15-1075} {A large annotated
  corpus for learning natural language inference}.
\newblock In \emph{Proceedings of the 2015 Conference on Empirical Methods in
  Natural Language Processing}, pages 632--642, Lisbon, Portugal. Association
  for Computational Linguistics.

\bibitem[{Chowdhery et~al.(2022)Chowdhery, Narang, Devlin, Bosma, Mishra,
  Roberts, Barham, Chung, Sutton, Gehrmann, Schuh, Shi, Tsvyashchenko, Maynez,
  Rao, Barnes, Tay, Shazeer, Prabhakaran, Reif, Du, Hutchinson, Pope, Bradbury,
  Austin, Isard, Gur-Ari, Yin, Duke, Levskaya, Ghemawat, Dev, Michalewski,
  Garcia, Misra, Robinson, Fedus, Zhou, Ippolito, Luan, Lim, Zoph, Spiridonov,
  Sepassi, Dohan, Agrawal, Omernick, Dai, Pillai, Pellat, Lewkowycz, Moreira,
  Child, Polozov, Lee, Zhou, Wang, Saeta, Diaz, Firat, Catasta, Wei,
  Meier-Hellstern, Eck, Dean, Petrov, and Fiedel}]{palm-2022}
Aakanksha Chowdhery, Sharan Narang, Jacob Devlin, Maarten Bosma, Gaurav Mishra,
  Adam Roberts, Paul Barham, Hyung~Won Chung, Charles Sutton, Sebastian
  Gehrmann, Parker Schuh, Kensen Shi, Sasha Tsvyashchenko, Joshua Maynez,
  Abhishek Rao, Parker Barnes, Yi~Tay, Noam Shazeer, Vinodkumar Prabhakaran,
  Emily Reif, Nan Du, Ben Hutchinson, Reiner Pope, James Bradbury, Jacob
  Austin, Michael Isard, Guy Gur-Ari, Pengcheng Yin, Toju Duke, Anselm
  Levskaya, Sanjay Ghemawat, Sunipa Dev, Henryk Michalewski, Xavier Garcia,
  Vedant Misra, Kevin Robinson, Liam Fedus, Denny Zhou, Daphne Ippolito, David
  Luan, Hyeontaek Lim, Barret Zoph, Alexander Spiridonov, Ryan Sepassi, David
  Dohan, Shivani Agrawal, Mark Omernick, Andrew~M. Dai,
  Thanumalayan~Sankaranarayana Pillai, Marie Pellat, Aitor Lewkowycz, Erica
  Moreira, Rewon Child, Oleksandr Polozov, Katherine Lee, Zongwei Zhou, Xuezhi
  Wang, Brennan Saeta, Mark Diaz, Orhan Firat, Michele Catasta, Jason Wei,
  Kathy Meier-Hellstern, Douglas Eck, Jeff Dean, Slav Petrov, and Noah Fiedel.
  2022.
\newblock \href {http://arxiv.org/abs/2204.02311} {Pa{LM}: Scaling language
  modeling with pathways}.
\newblock \emph{arXiv preprint arXiv:2204.02311}.

\bibitem[{Conneau et~al.(2018)Conneau, Rinott, Lample, Williams, Bowman,
  Schwenk, and Stoyanov}]{conneau-etal-2018-xnli}
Alexis Conneau, Ruty Rinott, Guillaume Lample, Adina Williams, Samuel Bowman,
  Holger Schwenk, and Veselin Stoyanov. 2018.
\newblock \href {https://doi.org/10.18653/v1/D18-1269} {{XNLI}: Evaluating
  cross-lingual sentence representations}.
\newblock In \emph{Proceedings of the 2018 Conference on Empirical Methods in
  Natural Language Processing}, pages 2475--2485, Brussels, Belgium.
  Association for Computational Linguistics.

\bibitem[{Dagan et~al.(2006)Dagan, Glickman, and Magnini}]{dagan2006pascal}
Ido Dagan, Oren Glickman, and Bernardo Magnini. 2006.
\newblock \href {https://link.springer.com/chapter/10.1007/11736790_9} {The
  {PASCAL} recognising textual entailment challenge}.
\newblock In \emph{Machine Learning Challenges. Evaluating Predictive
  Uncertainty, Visual Object Classification, and Recognising Tectual
  Entailment: First PASCAL Machine Learning Challenges Workshop, MLCW 2005,
  Southampton, UK, April 11-13, 2005, Revised Selected Papers}, pages 177--190.
  Springer.

\bibitem[{Dziri et~al.(2022)Dziri, Rashkin, Linzen, and
  Reitter}]{dziri-etal-2022-evaluating}
Nouha Dziri, Hannah Rashkin, Tal Linzen, and David Reitter. 2022.
\newblock \href {https://doi.org/10.1162/tacl_a_00506} {Evaluating attribution
  in dialogue systems: The {BEGIN} benchmark}.
\newblock \emph{Transactions of the Association for Computational Linguistics},
  10:1066--1083.

\bibitem[{Fabbri et~al.(2021)Fabbri, Kry{\'s}ci{\'n}ski, McCann, Xiong, Socher,
  and Radev}]{fabbri-etal-2021-summeval}
Alexander~R. Fabbri, Wojciech Kry{\'s}ci{\'n}ski, Bryan McCann, Caiming Xiong,
  Richard Socher, and Dragomir Radev. 2021.
\newblock \href {https://doi.org/10.1162/tacl_a_00373} {{S}umm{E}val:
  Re-evaluating summarization evaluation}.
\newblock \emph{Transactions of the Association for Computational Linguistics},
  9:391--409.

\bibitem[{Freitag et~al.(2021{\natexlab{a}})Freitag, Foster, Grangier,
  Ratnakar, Tan, and Macherey}]{freitag-etal-2021-experts}
Markus Freitag, George Foster, David Grangier, Viresh Ratnakar, Qijun Tan, and
  Wolfgang Macherey. 2021{\natexlab{a}}.
\newblock \href {https://doi.org/10.1162/tacl_a_00437} {Experts, errors, and
  context: A large-scale study of human evaluation for machine translation}.
\newblock \emph{Transactions of the Association for Computational Linguistics},
  9:1460--1474.

\bibitem[{Freitag et~al.(2021{\natexlab{b}})Freitag, Rei, Mathur, Lo, Stewart,
  Foster, Lavie, and Bojar}]{freitag-etal-2021-results}
Markus Freitag, Ricardo Rei, Nitika Mathur, Chi-kiu Lo, Craig Stewart, George
  Foster, Alon Lavie, and Ond{\v{r}}ej Bojar. 2021{\natexlab{b}}.
\newblock \href {https://aclanthology.org/2021.wmt-1.73} {Results of the
  {WMT}21 metrics shared task: Evaluating metrics with expert-based human
  evaluations on {TED} and news domain}.
\newblock In \emph{Proceedings of the Sixth Conference on Machine Translation},
  pages 733--774, Online. Association for Computational Linguistics.

\bibitem[{Fyodorov et~al.(2000)Fyodorov, Winter, and
  Francez}]{fyodorov2000natural}
Yaroslav Fyodorov, Yoad Winter, and Nissim Francez. 2000.
\newblock A natural logic inference system.
\newblock In \emph{Proceedings of the 2nd Workshop on Inference in
  Computational Semantics (ICoS-2)}.

\bibitem[{Gehrmann et~al.(2021)Gehrmann, Adewumi, Aggarwal, Ammanamanchi,
  Aremu, Bosselut, Chandu, Clinciu, Das, Dhole, Du, Durmus, Du{\v{s}}ek,
  Emezue, Gangal, Garbacea, Hashimoto, Hou, Jernite, Jhamtani, Ji, Jolly, Kale,
  Kumar, Ladhak, Madaan, Maddela, Mahajan, Mahamood, Majumder, Martins,
  McMillan-Major, Mille, van Miltenburg, Nadeem, Narayan, Nikolaev,
  Niyongabo~Rubungo, Osei, Parikh, Perez-Beltrachini, Rao, Raunak, Rodriguez,
  Santhanam, Sedoc, Sellam, Shaikh, Shimorina, Sobrevilla~Cabezudo, Strobelt,
  Subramani, Xu, Yang, Yerukola, and Zhou}]{gehrmann-etal-2021-gem}
Sebastian Gehrmann, Tosin Adewumi, Karmanya Aggarwal, Pawan~Sasanka
  Ammanamanchi, Anuoluwapo Aremu, Antoine Bosselut, Khyathi~Raghavi Chandu,
  Miruna-Adriana Clinciu, Dipanjan Das, Kaustubh Dhole, Wanyu Du, Esin Durmus,
  Ond{\v{r}}ej Du{\v{s}}ek, Chris~Chinenye Emezue, Varun Gangal, Cristina
  Garbacea, Tatsunori Hashimoto, Yufang Hou, Yacine Jernite, Harsh Jhamtani,
  Yangfeng Ji, Shailza Jolly, Mihir Kale, Dhruv Kumar, Faisal Ladhak, Aman
  Madaan, Mounica Maddela, Khyati Mahajan, Saad Mahamood, Bodhisattwa~Prasad
  Majumder, Pedro~Henrique Martins, Angelina McMillan-Major, Simon Mille, Emiel
  van Miltenburg, Moin Nadeem, Shashi Narayan, Vitaly Nikolaev, Andre
  Niyongabo~Rubungo, Salomey Osei, Ankur Parikh, Laura Perez-Beltrachini,
  Niranjan~Ramesh Rao, Vikas Raunak, Juan~Diego Rodriguez, Sashank Santhanam,
  Jo{\~a}o Sedoc, Thibault Sellam, Samira Shaikh, Anastasia Shimorina,
  Marco~Antonio Sobrevilla~Cabezudo, Hendrik Strobelt, Nishant Subramani, Wei
  Xu, Diyi Yang, Akhila Yerukola, and Jiawei Zhou. 2021.
\newblock \href {https://doi.org/10.18653/v1/2021.gem-1.10} {The {GEM}
  benchmark: Natural language generation, its evaluation and metrics}.
\newblock In \emph{Proceedings of the 1st Workshop on Natural Language
  Generation, Evaluation, and Metrics (GEM 2021)}, pages 96--120, Online.
  Association for Computational Linguistics.

\bibitem[{Gehrmann et~al.(2022)Gehrmann, Clark, and Sellam}]{repairing-2022}
Sebastian Gehrmann, Elizabeth Clark, and Thibault Sellam. 2022.
\newblock \href {https://doi.org/10.48550/ARXIV.2202.06935} {Repairing the
  cracked foundation: A survey of obstacles in evaluation practices for
  generated text}.
\newblock \emph{arXiv preprint arXiv:2202.06935}.

\bibitem[{Gekhman et~al.(2023)Gekhman, Herzig, Aharoni, Elkind, and
  Szpektor}]{gekhman2023trueteacher}
Zorik Gekhman, Jonathan Herzig, Roee Aharoni, Chen Elkind, and Idan Szpektor.
  2023.
\newblock \href {http://arxiv.org/abs/2305.11171} {True{T}eacher: Learning
  factual consistency evaluation with large language models}.
\newblock \emph{arXiv preprint arXiv:2305.11171}.

\bibitem[{Gupta et~al.(2022)Gupta, Wu, Liu, and
  Xiong}]{gupta-etal-2022-dialfact}
Prakhar Gupta, Chien-Sheng Wu, Wenhao Liu, and Caiming Xiong. 2022.
\newblock \href {https://doi.org/10.18653/v1/2022.acl-long.263} {{D}ial{F}act:
  A benchmark for fact-checking in dialogue}.
\newblock In \emph{Proceedings of the 60th Annual Meeting of the Association
  for Computational Linguistics (Volume 1: Long Papers)}, pages 3785--3801,
  Dublin, Ireland. Association for Computational Linguistics.

\bibitem[{Hasan et~al.(2021)Hasan, Bhattacharjee, Islam, Mubasshir, Li, Kang,
  Rahman, and Shahriyar}]{hasan-etal-2021-xl}
Tahmid Hasan, Abhik Bhattacharjee, Md.~Saiful Islam, Kazi Mubasshir, Yuan-Fang
  Li, Yong-Bin Kang, M.~Sohel Rahman, and Rifat Shahriyar. 2021.
\newblock \href {https://doi.org/10.18653/v1/2021.findings-acl.413} {{XL}-sum:
  Large-scale multilingual abstractive summarization for 44 languages}.
\newblock In \emph{Findings of the Association for Computational Linguistics:
  ACL-IJCNLP 2021}, pages 4693--4703, Online. Association for Computational
  Linguistics.

\bibitem[{Hermann et~al.(2015)Hermann, Kocisky, Grefenstette, Espeholt, Kay,
  Suleyman, and Blunsom}]{hermann2015teaching}
Karl~Moritz Hermann, Tomas Kocisky, Edward Grefenstette, Lasse Espeholt, Will
  Kay, Mustafa Suleyman, and Phil Blunsom. 2015.
\newblock \href
  {http://papers.neurips.cc/paper/5945-teaching-machines-to-read-and-comprehend.pdf}
  {Teaching machines to read and comprehend}.
\newblock \emph{Advances in Neural Information Processing Systems}, 28.

\bibitem[{Honovich et~al.(2022)Honovich, Aharoni, Herzig, Taitelbaum,
  Kukliansy, Cohen, Scialom, Szpektor, Hassidim, and
  Matias}]{honovich-etal-2022-true}
Or~Honovich, Roee Aharoni, Jonathan Herzig, Hagai Taitelbaum, Doron Kukliansy,
  Vered Cohen, Thomas Scialom, Idan Szpektor, Avinatan Hassidim, and Yossi
  Matias. 2022.
\newblock \href {https://doi.org/10.18653/v1/2022.dialdoc-1.19} {{TRUE}:
  Re-evaluating factual consistency evaluation}.
\newblock In \emph{Proceedings of the Second DialDoc Workshop on
  Document-grounded Dialogue and Conversational Question Answering}, pages
  161--175, Dublin, Ireland. Association for Computational Linguistics.

\bibitem[{Honovich et~al.(2021)Honovich, Choshen, Aharoni, Neeman, Szpektor,
  and Abend}]{honovich-etal-2021-q2}
Or~Honovich, Leshem Choshen, Roee Aharoni, Ella Neeman, Idan Szpektor, and Omri
  Abend. 2021.
\newblock \href {https://doi.org/10.18653/v1/2021.emnlp-main.619} {$q^{2}$:
  {E}valuating factual consistency in knowledge-grounded dialogues via question
  generation and question answering}.
\newblock In \emph{Proceedings of the 2021 Conference on Empirical Methods in
  Natural Language Processing}, pages 7856--7870, Online and Punta Cana,
  Dominican Republic. Association for Computational Linguistics.

\bibitem[{Ji et~al.(2023)Ji, Lee, Frieske, Yu, Su, Xu, Ishii, Bang, Madotto,
  and Fung}]{ji2023survey}
Ziwei Ji, Nayeon Lee, Rita Frieske, Tiezheng Yu, Dan Su, Yan Xu, Etsuko Ishii,
  Ye~Jin Bang, Andrea Madotto, and Pascale Fung. 2023.
\newblock \href {https://dl.acm.org/doi/10.1145/3571730} {Survey of
  hallucination in natural language generation}.
\newblock \emph{ACM Computing Surveys}, 55(12):1--38.

\bibitem[{Khashabi et~al.(2022)Khashabi, Lyu, Min, Qin, Richardson, Welleck,
  Hajishirzi, Khot, Sabharwal, Singh, and Choi}]{khashabi-etal-2022-prompt}
Daniel Khashabi, Xinxi Lyu, Sewon Min, Lianhui Qin, Kyle Richardson, Sean
  Welleck, Hannaneh Hajishirzi, Tushar Khot, Ashish Sabharwal, Sameer Singh,
  and Yejin Choi. 2022.
\newblock \href {https://doi.org/10.18653/v1/2022.naacl-main.266} {Prompt
  waywardness: The curious case of discretized interpretation of continuous
  prompts}.
\newblock In \emph{Proceedings of the 2022 Conference of the North American
  Chapter of the Association for Computational Linguistics: Human Language
  Technologies}, pages 3631--3643, Seattle, United States. Association for
  Computational Linguistics.

\bibitem[{Khot et~al.(2018)Khot, Sabharwal, and Clark}]{khot2018scitail}
Tushar Khot, Ashish Sabharwal, and Peter Clark. 2018.
\newblock \href {https://ojs.aaai.org/index.php/AAAI/article/view/12022}
  {Sci{t}ai{L}: A textual entailment dataset from science question answering}.
\newblock In \emph{Proceedings of the AAAI Conference on Artificial
  Intelligence}, volume~32.

\bibitem[{Kocmi et~al.(2021)Kocmi, Federmann, Grundkiewicz, Junczys-Dowmunt,
  Matsushita, and Menezes}]{kocmi-etal-2021-ship}
Tom Kocmi, Christian Federmann, Roman Grundkiewicz, Marcin Junczys-Dowmunt,
  Hitokazu Matsushita, and Arul Menezes. 2021.
\newblock \href {https://aclanthology.org/2021.wmt-1.57} {To ship or not to
  ship: An extensive evaluation of automatic metrics for machine translation}.
\newblock In \emph{Proceedings of the Sixth Conference on Machine Translation},
  pages 478--494, Online. Association for Computational Linguistics.

\bibitem[{Krippendorff(1980)}]{Krippendorff1980ContentAA}
Klaus Krippendorff. 1980.
\newblock Content analysis: An introduction to its methodology.

\bibitem[{Kryscinski et~al.(2020)Kryscinski, McCann, Xiong, and
  Socher}]{kryscinski-etal-2020-evaluating}
Wojciech Kryscinski, Bryan McCann, Caiming Xiong, and Richard Socher. 2020.
\newblock \href {https://doi.org/10.18653/v1/2020.emnlp-main.750} {Evaluating
  the factual consistency of abstractive text summarization}.
\newblock In \emph{Proceedings of the 2020 Conference on Empirical Methods in
  Natural Language Processing (EMNLP)}, pages 9332--9346, Online. Association
  for Computational Linguistics.

\bibitem[{Ladhak et~al.(2020)Ladhak, Durmus, Cardie, and
  McKeown}]{ladhak-etal-2020-wikilingua}
Faisal Ladhak, Esin Durmus, Claire Cardie, and Kathleen McKeown. 2020.
\newblock \href {https://doi.org/10.18653/v1/2020.findings-emnlp.360}
  {{W}iki{L}ingua: A new benchmark dataset for cross-lingual abstractive
  summarization}.
\newblock In \emph{Findings of the Association for Computational Linguistics:
  EMNLP 2020}, pages 4034--4048, Online. Association for Computational
  Linguistics.

\bibitem[{Lin(2004)}]{lin-2004-rouge}
Chin-Yew Lin. 2004.
\newblock \href {https://aclanthology.org/W04-1013} {{ROUGE}: A package for
  automatic evaluation of summaries}.
\newblock In \emph{Text Summarization Branches Out}, pages 74--81, Barcelona,
  Spain. Association for Computational Linguistics.

\bibitem[{Liu et~al.(2023)Liu, Iter, Xu, Wang, Xu, and Zhu}]{gpteval-2023}
Yang Liu, Dan Iter, Yichong Xu, Shuohang Wang, Ruochen Xu, and Chenguang Zhu.
  2023.
\newblock \href {http://arxiv.org/abs/2303.16634} {G-{E}val: {NLG} evaluation
  using {GPT}-4 with better human alignment}.
\newblock \emph{arXiv preprint arXiv:2303.16634}.

\bibitem[{Liu et~al.(2022)Liu, Fabbri, Liu, Zhao, Nan, Han, Han, Joty, Wu,
  Xiong, and Radev}]{liu2022revisiting}
Yixin Liu, Alexander~R. Fabbri, Pengfei Liu, Yilun Zhao, Linyong Nan, Ruilin
  Han, Simeng Han, Shafiq Joty, Chien-Sheng Wu, Caiming Xiong, and Dragomir
  Radev. 2022.
\newblock \href {http://arxiv.org/abs/2212.07981} {Revisiting the gold
  standard: Grounding summarization evaluation with robust human evaluation}.
\newblock \emph{arXiv preprint arXiv:2212.07981}.

\bibitem[{Maynez et~al.(2020)Maynez, Narayan, Bohnet, and
  McDonald}]{maynez-etal-2020-faithfulness}
Joshua Maynez, Shashi Narayan, Bernd Bohnet, and Ryan McDonald. 2020.
\newblock \href {https://doi.org/10.18653/v1/2020.acl-main.173} {On
  faithfulness and factuality in abstractive summarization}.
\newblock In \emph{Proceedings of the 58th Annual Meeting of the Association
  for Computational Linguistics}, pages 1906--1919, Online. Association for
  Computational Linguistics.

\bibitem[{Narayan et~al.(2018)Narayan, Cohen, and
  Lapata}]{narayan-etal-2018-dont}
Shashi Narayan, Shay~B. Cohen, and Mirella Lapata. 2018.
\newblock \href {https://doi.org/10.18653/v1/D18-1206} {Don{'}t give me the
  details, just the summary! topic-aware convolutional neural networks for
  extreme summarization}.
\newblock In \emph{Proceedings of the 2018 Conference on Empirical Methods in
  Natural Language Processing}, pages 1797--1807, Brussels, Belgium.
  Association for Computational Linguistics.

\bibitem[{Ouyang et~al.(2022)Ouyang, Wu, Jiang, Almeida, Wainwright, Mishkin,
  Zhang, Agarwal, Slama, Ray, Schulman, Hilton, Kelton, Miller, Simens, Askell,
  Welinder, Christiano, Leike, and Lowe}]{instruct-gpt-2022}
Long Ouyang, Jeff Wu, Xu~Jiang, Diogo Almeida, Carroll~L. Wainwright, Pamela
  Mishkin, Chong Zhang, Sandhini Agarwal, Katarina Slama, Alex Ray, John
  Schulman, Jacob Hilton, Fraser Kelton, Luke Miller, Maddie Simens, Amanda
  Askell, Peter Welinder, Paul Christiano, Jan Leike, and Ryan Lowe. 2022.
\newblock \href {https://doi.org/10.48550/ARXIV.2203.02155} {Training language
  models to follow instructions with human feedback}.
\newblock \emph{arXiv preprint arXiv:2203.02155}.

\bibitem[{Pagnoni et~al.(2021)Pagnoni, Balachandran, and
  Tsvetkov}]{pagnoni-etal-2021-understanding}
Artidoro Pagnoni, Vidhisha Balachandran, and Yulia Tsvetkov. 2021.
\newblock \href {https://doi.org/10.18653/v1/2021.naacl-main.383}
  {Understanding factuality in abstractive summarization with {FRANK}: A
  benchmark for factuality metrics}.
\newblock In \emph{Proceedings of the 2021 Conference of the North American
  Chapter of the Association for Computational Linguistics: Human Language
  Technologies}, pages 4812--4829, Online. Association for Computational
  Linguistics.

\bibitem[{Papineni et~al.(2002)Papineni, Roukos, Ward, and
  Zhu}]{papineni-etal-2002-bleu}
Kishore Papineni, Salim Roukos, Todd Ward, and Wei-Jing Zhu. 2002.
\newblock \href {https://doi.org/10.3115/1073083.1073135} {{B}leu: a method for
  automatic evaluation of machine translation}.
\newblock In \emph{Proceedings of the 40th Annual Meeting of the Association
  for Computational Linguistics}, pages 311--318, Philadelphia, Pennsylvania,
  USA. Association for Computational Linguistics.

\bibitem[{Parikh et~al.(2020)Parikh, Wang, Gehrmann, Faruqui, Dhingra, Yang,
  and Das}]{parikh-etal-2020-totto}
Ankur Parikh, Xuezhi Wang, Sebastian Gehrmann, Manaal Faruqui, Bhuwan Dhingra,
  Diyi Yang, and Dipanjan Das. 2020.
\newblock \href {https://doi.org/10.18653/v1/2020.emnlp-main.89} {{ToTTo}: A
  controlled table-to-text generation dataset}.
\newblock In \emph{Proceedings of the 2020 Conference on Empirical Methods in
  Natural Language Processing (EMNLP)}, pages 1173--1186, Online. Association
  for Computational Linguistics.

\bibitem[{Radford et~al.(2019)Radford, Wu, Child, Luan, Amodei, Sutskever
  et~al.}]{radford2019language}
Alec Radford, Jeffrey Wu, Rewon Child, David Luan, Dario Amodei, Ilya
  Sutskever, et~al. 2019.
\newblock \href
  {https://d4mucfpksywv.cloudfront.net/better-language-models/language_models_are_unsupervised_multitask_learners.pdf}
  {Language models are unsupervised multitask learners}.
\newblock \emph{OpenAI blog}, 1(8):9.

\bibitem[{Raffel et~al.(2020)Raffel, Shazeer, Roberts, Lee, Narang, Matena,
  Zhou, Li, and Liu}]{t5-2020}
Colin Raffel, Noam Shazeer, Adam Roberts, Katherine Lee, Sharan Narang, Michael
  Matena, Yanqi Zhou, Wei Li, and Peter~J. Liu. 2020.
\newblock \href {http://jmlr.org/papers/v21/20-074.html} {Exploring the limits
  of transfer learning with a unified text-to-text transformer}.
\newblock \emph{Journal of Machine Learning Research}, 21(140):1--67.

\bibitem[{Rashkin et~al.(2021)Rashkin, Nikolaev, Lamm, Aroyo, Collins, Das,
  Petrov, Tomar, Turc, and Reitter}]{rashkin2021measuring}
Hannah Rashkin, Vitaly Nikolaev, Matthew Lamm, Lora Aroyo, Michael Collins,
  Dipanjan Das, Slav Petrov, Gaurav~Singh Tomar, Iulia Turc, and David Reitter.
  2021.
\newblock \href {https://arxiv.org/abs/2112.12870} {Measuring attribution in
  natural language generation models}.
\newblock \emph{arXiv preprint arXiv:2112.12870}.

\bibitem[{Rei et~al.(2020)Rei, Stewart, Farinha, and
  Lavie}]{rei-etal-2020-comet}
Ricardo Rei, Craig Stewart, Ana~C Farinha, and Alon Lavie. 2020.
\newblock \href {https://doi.org/10.18653/v1/2020.emnlp-main.213} {{COMET}: A
  neural framework for {MT} evaluation}.
\newblock In \emph{Proceedings of the 2020 Conference on Empirical Methods in
  Natural Language Processing (EMNLP)}, pages 2685--2702, Online. Association
  for Computational Linguistics.

\bibitem[{Roberts et~al.(2022)Roberts, Chung, Levskaya, Mishra, Bradbury,
  Andor, Narang, Lester, Gaffney, Mohiuddin et~al.}]{roberts2022scaling}
Adam Roberts, Hyung~Won Chung, Anselm Levskaya, Gaurav Mishra, James Bradbury,
  Daniel Andor, Sharan Narang, Brian Lester, Colin Gaffney, Afroz Mohiuddin,
  et~al. 2022.
\newblock \href {https://arxiv.org/abs/2203.17189} {Scaling up models and data
  with t5x and seqio}.
\newblock \emph{arXiv preprint arXiv:2203.17189}.

\bibitem[{Schuster et~al.(2021)Schuster, Fisch, and
  Barzilay}]{schuster-etal-2021-get}
Tal Schuster, Adam Fisch, and Regina Barzilay. 2021.
\newblock \href {https://doi.org/10.18653/v1/2021.naacl-main.52} {Get your
  vitamin {C}! robust fact verification with contrastive evidence}.
\newblock In \emph{Proceedings of the 2021 Conference of the North American
  Chapter of the Association for Computational Linguistics: Human Language
  Technologies}, pages 624--643, Online. Association for Computational
  Linguistics.

\bibitem[{Scialom et~al.(2020)Scialom, Dray, Lamprier, Piwowarski, and
  Staiano}]{scialom-etal-2020-mlsum}
Thomas Scialom, Paul-Alexis Dray, Sylvain Lamprier, Benjamin Piwowarski, and
  Jacopo Staiano. 2020.
\newblock \href {https://doi.org/10.18653/v1/2020.emnlp-main.647} {{MLSUM}: The
  multilingual summarization corpus}.
\newblock In \emph{Proceedings of the 2020 Conference on Empirical Methods in
  Natural Language Processing (EMNLP)}, pages 8051--8067, Online. Association
  for Computational Linguistics.

\bibitem[{Sellam et~al.(2020)Sellam, Das, and Parikh}]{sellam-etal-2020-bleurt}
Thibault Sellam, Dipanjan Das, and Ankur Parikh. 2020.
\newblock \href {https://doi.org/10.18653/v1/2020.acl-main.704} {{BLEURT}:
  Learning robust metrics for text generation}.
\newblock In \emph{Proceedings of the 58th Annual Meeting of the Association
  for Computational Linguistics}, pages 7881--7892, Online. Association for
  Computational Linguistics.

\bibitem[{Stiennon et~al.(2020)Stiennon, Ouyang, Wu, Ziegler, Lowe, Voss,
  Radford, Amodei, and Christiano}]{stiennon2020learning}
Nisan Stiennon, Long Ouyang, Jeffrey Wu, Daniel Ziegler, Ryan Lowe, Chelsea
  Voss, Alec Radford, Dario Amodei, and Paul~F Christiano. 2020.
\newblock \href
  {https://proceedings.neurips.cc/paper_files/paper/2020/file/1f89885d556929e98d3ef9b86448f951-Paper.pdf}
  {Learning to summarize with human feedback}.
\newblock In \emph{Advances in Neural Information Processing Systems},
  volume~33, pages 3008--3021. Curran Associates, Inc.

\bibitem[{Thorne et~al.(2018)Thorne, Vlachos, Christodoulopoulos, and
  Mittal}]{thorne-etal-2018-fever}
James Thorne, Andreas Vlachos, Christos Christodoulopoulos, and Arpit Mittal.
  2018.
\newblock \href {https://doi.org/10.18653/v1/N18-1074} {{FEVER}: a large-scale
  dataset for fact extraction and {VER}ification}.
\newblock In \emph{Proceedings of the 2018 Conference of the North {A}merican
  Chapter of the Association for Computational Linguistics: Human Language
  Technologies, Volume 1 (Long Papers)}, pages 809--819, New Orleans,
  Louisiana. Association for Computational Linguistics.

\bibitem[{Tian et~al.(2019)Tian, Narayan, Sellam, and
  Parikh}]{tian2019sticking}
Ran Tian, Shashi Narayan, Thibault Sellam, and Ankur~P Parikh. 2019.
\newblock \href {https://arxiv.org/abs/1910.08684} {Sticking to the facts:
  Confident decoding for faithful data-to-text generation}.
\newblock \emph{arXiv preprint arXiv:1910.08684}.

\bibitem[{V{\"o}lske et~al.(2017)V{\"o}lske, Potthast, Syed, and
  Stein}]{volske-etal-2017-tl}
Michael V{\"o}lske, Martin Potthast, Shahbaz Syed, and Benno Stein. 2017.
\newblock \href {https://doi.org/10.18653/v1/W17-4508} {{TL};{DR}: Mining
  {R}eddit to learn automatic summarization}.
\newblock In \emph{Proceedings of the Workshop on New Frontiers in
  Summarization}, pages 59--63, Copenhagen, Denmark. Association for
  Computational Linguistics.

\bibitem[{Wang et~al.(2020)Wang, Cho, and Lewis}]{wang-etal-2020-asking}
Alex Wang, Kyunghyun Cho, and Mike Lewis. 2020.
\newblock \href {https://doi.org/10.18653/v1/2020.acl-main.450} {Asking and
  answering questions to evaluate the factual consistency of summaries}.
\newblock In \emph{Proceedings of the 58th Annual Meeting of the Association
  for Computational Linguistics}, pages 5008--5020, Online. Association for
  Computational Linguistics.

\bibitem[{Williams et~al.(2018)Williams, Nangia, and
  Bowman}]{williams-etal-2018-broad}
Adina Williams, Nikita Nangia, and Samuel Bowman. 2018.
\newblock \href {https://doi.org/10.18653/v1/N18-1101} {A broad-coverage
  challenge corpus for sentence understanding through inference}.
\newblock In \emph{Proceedings of the 2018 Conference of the North {A}merican
  Chapter of the Association for Computational Linguistics: Human Language
  Technologies, Volume 1 (Long Papers)}, pages 1112--1122, New Orleans,
  Louisiana. Association for Computational Linguistics.

\bibitem[{Wiseman et~al.(2017)Wiseman, Shieber, and
  Rush}]{wiseman-etal-2017-challenges}
Sam Wiseman, Stuart Shieber, and Alexander Rush. 2017.
\newblock \href {https://doi.org/10.18653/v1/D17-1239} {Challenges in
  data-to-document generation}.
\newblock In \emph{Proceedings of the 2017 Conference on Empirical Methods in
  Natural Language Processing}, pages 2253--2263, Copenhagen, Denmark.
  Association for Computational Linguistics.

\bibitem[{Xue et~al.(2021)Xue, Constant, Roberts, Kale, Al-Rfou, Siddhant,
  Barua, and Raffel}]{xue-etal-2021-mt5}
Linting Xue, Noah Constant, Adam Roberts, Mihir Kale, Rami Al-Rfou, Aditya
  Siddhant, Aditya Barua, and Colin Raffel. 2021.
\newblock \href {https://doi.org/10.18653/v1/2021.naacl-main.41} {m{T}5: A
  massively multilingual pre-trained text-to-text transformer}.
\newblock In \emph{Proceedings of the 2021 Conference of the North American
  Chapter of the Association for Computational Linguistics: Human Language
  Technologies}, pages 483--498, Online. Association for Computational
  Linguistics.

\bibitem[{Zhang* et~al.(2020)Zhang*, Kishore*, Wu*, Weinberger, and
  Artzi}]{bert-score}
Tianyi Zhang*, Varsha Kishore*, Felix Wu*, Kilian~Q. Weinberger, and Yoav
  Artzi. 2020.
\newblock \href {https://openreview.net/forum?id=SkeHuCVFDr} {{BERTS}core:
  Evaluating text generation with bert}.
\newblock In \emph{International Conference on Learning Representations}.

\bibitem[{Zhang et~al.(2019)Zhang, Baldridge, and He}]{zhang-etal-2019-paws}
Yuan Zhang, Jason Baldridge, and Luheng He. 2019.
\newblock \href {https://doi.org/10.18653/v1/N19-1131} {{PAWS}: Paraphrase
  adversaries from word scrambling}.
\newblock In \emph{Proceedings of the 2019 Conference of the North {A}merican
  Chapter of the Association for Computational Linguistics: Human Language
  Technologies, Volume 1 (Long and Short Papers)}, pages 1298--1308,
  Minneapolis, Minnesota. Association for Computational Linguistics.

\bibitem[{Zhou et~al.(2021)Zhou, Neubig, Gu, Diab, Guzm{\'a}n, Zettlemoyer, and
  Ghazvininejad}]{zhou-etal-2021-detecting}
Chunting Zhou, Graham Neubig, Jiatao Gu, Mona Diab, Francisco Guzm{\'a}n, Luke
  Zettlemoyer, and Marjan Ghazvininejad. 2021.
\newblock \href {https://doi.org/10.18653/v1/2021.findings-acl.120} {Detecting
  hallucinated content in conditional neural sequence generation}.
\newblock In \emph{Findings of the Association for Computational Linguistics:
  ACL-IJCNLP 2021}, pages 1393--1404, Online. Association for Computational
  Linguistics.

\end{thebibliography}
\bibliographystyle{acl_natbib}

\appendix

\section{Training details}\label{app:training}
The summarization models were trained on the training split of each summarization dataset, with the exception of palm\_1shot, which generated a summary given a single example from the dataset and the input article.
The checkpoint for each model was selected using performance on the validation set of each respective dataset, except for t5\_base\_250 and mt5\_small\_250, which were only trained for 250 steps.
The input length for the T5 and mT5 models was set to 1024, and 2048 for PaLM. The target length was 512.

The \seahorse metrics were trained on the \seahorse training split, and the best checkpoint was selected based on performance on the validation set.
A separate metric was trained for each of the 6 dimensions of quality.
We used only ``Yes'' and ``No'' ratings for training and testing the \seahorse metrics.
The input length for the learnt metrics model is 2048.
The article and summary are separated with ``premise:'' and ``hypothesis:'' tags, respectively, to be consistent with \citet{honovich-etal-2022-true}.

All training and inference was done with the t5x framework \citep{roberts2022scaling} and run with TPU accelerators.

\section{Rate of positive responses}\label{app:yes_rate}
\autoref{tab:yes_rate} shows a detailed breakdown of the proportion of responses that were positive (i.e., ``Yes''), divided by language, dataset, model, and question.
Summaries in languages other than English and produced by smaller models tend to have lower scores, indicating good directions for improving our summarization systems.  

While most articles in the dataset were assigned to a subset of the summarization models, some articles were summarized by all 9 summarization systems (or 6 systems for the non-\en languages that did not use the T5 models).
Specifically in the test set, there were 543 articles that were summarized by all summarization systems.
\autoref{tab:test2_yes_rate} shows the positive response rate across those summaries.
\begin{table*}[htb]
\tiny
\begin{subtable}[t]{.5\textwidth}
\begin{tabular}[t]{llllllll}
\toprule
\multicolumn{8}{c}{DE ``YES'' RATE} \\ \hline
Dataset & Model & Q1 & Q2 & Q3 & Q4 & Q5 & Q6 \\ \hline
\multirow{7}{*}{mlsum} &
reference & 0.99 & 0.99 & 0.98 & 0.82 & 0.64 & 0.55 \\
& mt5\_small\_250 & 0.83 & 0.58 & 0.59 & 0.68 & 0.41 & 0.29 \\
& mt5\_small & 0.93 & 0.85 & 0.87 & 0.68 & 0.47 & 0.38 \\
& mt5\_xxl & 0.98 & 0.97 & 0.95 & 0.8 & 0.59 & 0.5 \\
& palm\_1shot & 0.93 & 0.93 & 0.9 & 0.83 & 0.73 & 0.66 \\
& palm\_finetuned & 0.99 & 0.99 & 0.99 & 0.88 & 0.82 & 0.73 \\
\cline{2-8} & total & 0.94 & 0.89 & 0.88 & 0.79 & 0.62 & 0.53 \\
\hline
\multirow{7}{*}{wikilingua} &
reference & 0.97 & 0.96 & 0.94 & 0.65 & 0.63 & 0.49 \\
& mt5\_small\_250 & 0.82 & 0.75 & 0.75 & 0.08 & 0.07 & 0.03 \\
& mt5\_small & 0.91 & 0.35 & 0.84 & 0.4 & 0.26 & 0.16 \\
& mt5\_xxl & 0.97 & 0.91 & 0.93 & 0.69 & 0.62 & 0.49 \\
& palm\_1shot & 0.76 & 0.72 & 0.73 & 0.63 & 0.53 & 0.42 \\
& palm\_finetuned & 0.98 & 0.97 & 0.95 & 0.74 & 0.79 & 0.65 \\
\cline{2-8} & total & 0.9 & 0.78 & 0.85 & 0.53 & 0.48 & 0.37 \\
\hline
total &  & 0.92 & 0.84 & 0.87 & 0.66 & 0.55 & 0.45 \\
\midrule
\multicolumn{8}{c}{EN ``YES'' RATE} \\ \hline
Dataset & Model & Q1 & Q2 & Q3 & Q4 & Q5 & Q6 \\ \hline
\multirow{10}{*}{xsum} &
reference & 1.0 & 1.0 & 0.96 & 0.54 & 0.68 & 0.47 \\
& t5\_base\_250 & 0.96 & 0.88 & 0.89 & 0.32 & 0.43 & 0.24 \\
& t5\_base & 0.96 & 0.91 & 0.91 & 0.42 & 0.5 & 0.32 \\
& t5\_xxl & 0.99 & 0.98 & 0.97 & 0.58 & 0.64 & 0.47 \\
& mt5\_small\_250 & 0.7 & 0.47 & 0.57 & 0.17 & 0.2 & 0.09 \\
& mt5\_small & 0.84 & 0.68 & 0.75 & 0.17 & 0.24 & 0.12 \\
& mt5\_xxl & 0.97 & 0.95 & 0.93 & 0.46 & 0.58 & 0.37 \\
& palm\_1shot & 0.97 & 0.96 & 0.91 & 0.48 & 0.55 & 0.39 \\
& palm\_finetuned & 0.99 & 0.99 & 0.99 & 0.6 & 0.65 & 0.51 \\
\cline{2-8} & total & 0.93 & 0.87 & 0.87 & 0.42 & 0.5 & 0.33 \\
\hline
\multirow{10}{*}{xlsum} &
reference & 1.0 & 1.0 & 0.97 & 0.6 & 0.74 & 0.51 \\
& t5\_base\_250 & 0.98 & 0.93 & 0.92 & 0.59 & 0.59 & 0.43 \\
& t5\_base & 0.99 & 0.96 & 0.96 & 0.65 & 0.68 & 0.52 \\
& t5\_xxl & 1.0 & 0.99 & 0.97 & 0.68 & 0.72 & 0.54 \\
& mt5\_small\_250 & 0.74 & 0.53 & 0.59 & 0.29 & 0.24 & 0.15 \\
& mt5\_small & 0.89 & 0.78 & 0.79 & 0.4 & 0.44 & 0.29 \\
& mt5\_xxl & 0.99 & 0.98 & 0.94 & 0.62 & 0.73 & 0.52 \\
& palm\_1shot & 0.95 & 0.95 & 0.92 & 0.73 & 0.68 & 0.58 \\
& palm\_finetuned & 1.0 & 1.0 & 1.0 & 0.62 & 0.59 & 0.45 \\
\cline{2-8} & total & 0.95 & 0.9 & 0.9 & 0.57 & 0.6 & 0.44 \\
\hline
\multirow{10}{*}{wikilingua} &
reference & 0.99 & 0.99 & 0.93 & 0.55 & 0.59 & 0.42 \\
& t5\_base\_250 & 0.98 & 0.59 & 0.93 & 0.31 & 0.26 & 0.09 \\
& t5\_base & 0.98 & 0.89 & 0.93 & 0.67 & 0.57 & 0.45 \\
& t5\_xxl & 0.98 & 0.95 & 0.92 & 0.68 & 0.63 & 0.51 \\
& mt5\_small\_250 & 0.96 & 0.27 & 0.91 & 0.45 & 0.09 & 0.02 \\
& mt5\_small & 0.95 & 0.65 & 0.88 & 0.52 & 0.37 & 0.19 \\
& mt5\_xxl & 1.0 & 0.96 & 0.92 & 0.62 & 0.64 & 0.49 \\
& palm\_1shot & 0.98 & 0.95 & 0.94 & 0.8 & 0.58 & 0.49 \\
& palm\_finetuned & 0.99 & 0.98 & 0.95 & 0.6 & 0.63 & 0.54 \\
\cline{2-8} & total & 0.98 & 0.8 & 0.92 & 0.58 & 0.48 & 0.35 \\
\hline
total &  & 0.95 & 0.86 & 0.9 & 0.53 & 0.53 & 0.38 \\
\midrule
\multicolumn{8}{c}{ES ``YES'' RATE} \\ \hline
Dataset & Model & Q1 & Q2 & Q3 & Q4 & Q5 & Q6 \\ \hline
\multirow{7}{*}{mlsum} &
reference & 0.99 & 0.99 & 0.88 & 0.69 & 0.49 & 0.33 \\
& mt5\_small\_250 & 0.78 & 0.69 & 0.63 & 0.38 & 0.2 & 0.11 \\
& mt5\_small & 0.94 & 0.88 & 0.8 & 0.61 & 0.38 & 0.25 \\
& mt5\_xxl & 0.98 & 0.97 & 0.86 & 0.76 & 0.53 & 0.39 \\
& palm\_1shot & 0.73 & 0.72 & 0.27 & 0.41 & 0.45 & 0.32 \\
& palm\_finetuned & 0.99 & 0.99 & 0.02 & 0.92 & 0.78 & 0.75 \\
\cline{2-8} & total & 0.9 & 0.87 & 0.57 & 0.63 & 0.47 & 0.36 \\
\hline
\multirow{7}{*}{xlsum} &
reference & 0.99 & 0.99 & 0.96 & 0.31 & 0.49 & 0.21 \\
& mt5\_small\_250 & 0.64 & 0.44 & 0.55 & 0.17 & 0.16 & 0.07 \\
& mt5\_small & 0.8 & 0.63 & 0.71 & 0.23 & 0.28 & 0.12 \\
& mt5\_xxl & 0.98 & 0.96 & 0.94 & 0.39 & 0.43 & 0.23 \\
& palm\_1shot & 0.9 & 0.89 & 0.85 & 0.76 & 0.7 & 0.64 \\
& palm\_finetuned & 0.99 & 0.99 & 0.98 & 0.5 & 0.66 & 0.41 \\
\cline{2-8} & total & 0.88 & 0.81 & 0.83 & 0.39 & 0.45 & 0.28 \\
\hline
\multirow{7}{*}{wikilingua} &
reference & 0.99 & 0.97 & 0.96 & 0.5 & 0.62 & 0.35 \\
& mt5\_small\_250 & 0.75 & 0.61 & 0.73 & 0.16 & 0.08 & 0.03 \\
& mt5\_small & 0.95 & 0.37 & 0.92 & 0.42 & 0.28 & 0.11 \\
& mt5\_xxl & 0.98 & 0.93 & 0.95 & 0.57 & 0.64 & 0.4 \\
& palm\_1shot & 0.96 & 0.91 & 0.93 & 0.85 & 0.62 & 0.46 \\
& palm\_finetuned & 0.99 & 0.97 & 0.94 & 0.84 & 0.84 & 0.74 \\
\cline{2-8} & total & 0.93 & 0.79 & 0.9 & 0.55 & 0.51 & 0.34 \\
\hline
total &  & 0.91 & 0.83 & 0.77 & 0.52 & 0.48 & 0.33 \\
\bottomrule
\end{tabular}
\end{subtable}
\begin{subtable}[t]{.5\textwidth}
\begin{tabular}[t]{llllllll}
\toprule
\multicolumn{8}{c}{RU ``YES'' RATE} \\ \hline
Dataset & Model & Q1 & Q2 & Q3 & Q4 & Q5 & Q6 \\ \hline
\multirow{7}{*}{xlsum} &
reference & 0.99 & 0.98 & 0.94 & 0.48 & 0.82 & 0.44 \\
& mt5\_small\_250 & 0.4 & 0.21 & 0.29 & 0.2 & 0.25 & 0.1 \\
& mt5\_small & 0.73 & 0.58 & 0.57 & 0.27 & 0.47 & 0.19 \\
& mt5\_xxl & 0.95 & 0.93 & 0.83 & 0.44 & 0.76 & 0.4 \\
& palm\_1shot & 0.89 & 0.89 & 0.82 & 0.78 & 0.66 & 0.6 \\
& palm\_finetuned & 1.0 & 1.0 & 0.98 & 0.68 & 0.83 & 0.6 \\
\cline{2-8} & total & 0.83 & 0.77 & 0.74 & 0.48 & 0.64 & 0.39 \\
\hline
\multirow{7}{*}{wikilingua} &
reference & 0.97 & 0.95 & 0.9 & 0.56 & 0.65 & 0.46 \\
& mt5\_small\_250 & 0.73 & 0.22 & 0.66 & 0.31 & 0.05 & 0.04 \\
& mt5\_small & 0.83 & 0.26 & 0.75 & 0.39 & 0.17 & 0.09 \\
& mt5\_xxl & 0.96 & 0.92 & 0.85 & 0.54 & 0.61 & 0.45 \\
& palm\_1shot & 0.92 & 0.86 & 0.86 & 0.74 & 0.48 & 0.36 \\
& palm\_finetuned & 0.93 & 0.93 & 0.89 & 0.66 & 0.59 & 0.51 \\
\cline{2-8} & total & 0.89 & 0.69 & 0.82 & 0.53 & 0.42 & 0.32 \\
\hline
total &  & 0.86 & 0.73 & 0.78 & 0.5 & 0.53 & 0.35 \\
\midrule
\multicolumn{8}{c}{TR ``YES'' RATE} \\ \hline
Dataset & Model & Q1 & Q2 & Q3 & Q4 & Q5 & Q6 \\ \hline
\multirow{7}{*}{xlsum} &
reference & 1.0 & 1.0 & 0.88 & 0.46 & 0.82 & 0.43 \\
& mt5\_small\_250 & 0.59 & 0.41 & 0.34 & 0.23 & 0.33 & 0.17 \\
& mt5\_small & 0.85 & 0.72 & 0.57 & 0.35 & 0.49 & 0.29 \\
& mt5\_xxl & 0.99 & 0.98 & 0.83 & 0.54 & 0.78 & 0.49 \\
& palm\_1shot & 0.83 & 0.8 & 0.73 & 0.77 & 0.72 & 0.66 \\
& palm\_finetuned & 1.0 & 0.99 & 0.9 & 0.62 & 0.83 & 0.57 \\
\cline{2-8} & total & 0.87 & 0.81 & 0.7 & 0.48 & 0.65 & 0.42 \\
\hline
\multirow{7}{*}{wikilingua} &
reference & 0.94 & 0.92 & 0.83 & 0.5 & 0.73 & 0.46 \\
& mt5\_small\_250 & 0.9 & 0.34 & 0.79 & 0.35 & 0.2 & 0.12 \\
& mt5\_small & 0.82 & 0.53 & 0.57 & 0.1 & 0.18 & 0.05 \\
& mt5\_xxl & 0.93 & 0.89 & 0.77 & 0.44 & 0.61 & 0.35 \\
& palm\_1shot & 0.84 & 0.77 & 0.76 & 0.7 & 0.63 & 0.49 \\
& palm\_finetuned & 0.94 & 0.93 & 0.87 & 0.69 & 0.74 & 0.62 \\
\cline{2-8} & total & 0.89 & 0.72 & 0.76 & 0.44 & 0.5 & 0.33 \\
\hline
total &  & 0.88 & 0.78 & 0.72 & 0.47 & 0.61 & 0.39 \\
\midrule
\multicolumn{8}{c}{VI ``YES'' RATE} \\ \hline
Dataset & Model & Q1 & Q2 & Q3 & Q4 & Q5 & Q6 \\ \hline
\multirow{7}{*}{xlsum} &
reference & 0.86 & 0.85 & 0.81 & 0.37 & 0.65 & 0.35 \\
& mt5\_small\_250 & 0.49 & 0.33 & 0.39 & 0.09 & 0.17 & 0.06 \\
& mt5\_small & 0.7 & 0.57 & 0.59 & 0.2 & 0.41 & 0.15 \\
& mt5\_xxl & 0.84 & 0.83 & 0.8 & 0.38 & 0.67 & 0.36 \\
& palm\_1shot & 0.92 & 0.9 & 0.83 & 0.69 & 0.43 & 0.3 \\
& palm\_finetuned & 0.99 & 0.99 & 0.93 & 0.52 & 0.67 & 0.42 \\
\cline{2-8} & total & 0.8 & 0.75 & 0.73 & 0.37 & 0.51 & 0.28 \\
\hline
\multirow{7}{*}{wikilingua} &
reference & 0.98 & 0.97 & 0.94 & 0.57 & 0.71 & 0.51 \\
& mt5\_small\_250 & 0.82 & 0.28 & 0.78 & 0.25 & 0.1 & 0.06 \\
& mt5\_small & 0.91 & 0.28 & 0.87 & 0.31 & 0.25 & 0.13 \\
& mt5\_xxl & 0.97 & 0.95 & 0.93 & 0.49 & 0.65 & 0.42 \\
& palm\_1shot & 0.78 & 0.76 & 0.63 & 0.64 & 0.22 & 0.16 \\
& palm\_finetuned & 0.99 & 0.98 & 0.96 & 0.73 & 0.39 & 0.33 \\
\cline{2-8} & total & 0.91 & 0.7 & 0.86 & 0.49 & 0.39 & 0.27 \\
\hline
total &  & 0.85 & 0.72 & 0.79 & 0.43 & 0.45 & 0.27 \\
\bottomrule
\end{tabular}
\end{subtable}
\caption{The percent of ``Yes'' responses, broken down by language, dataset, model, and question number.}\label{tab:yes_rate}
\end{table*}
\begin{table*}[t]
\centering
\small
\begin{tabular}[t]{lllllll}
\toprule
Model & Q1 & Q2 & Q3 & Q4 & Q5 & Q6 \\ \hline
reference & 0.97 & 0.96 & 0.91 & 0.46 & 0.66 & 0.39 \\
t5\_base\_250 & 0.96 & 0.9 & 0.9 & 0.44 & 0.48 & 0.31 \\
t5\_base & 0.98 & 0.95 & 0.94 & 0.51 & 0.58 & 0.38 \\
t5\_xxl & 0.99 & 0.98 & 0.95 & 0.67 & 0.72 & 0.58 \\
mt5\_small\_250 & 0.64 & 0.45 & 0.5 & 0.26 & 0.24 & 0.12 \\
mt5\_small & 0.81 & 0.67 & 0.68 & 0.34 & 0.37 & 0.2 \\
mt5\_xxl & 0.95 & 0.94 & 0.89 & 0.5 & 0.66 & 0.37 \\
palm\_1shot & 0.93 & 0.88 & 0.86 & 0.75 & 0.56 & 0.44 \\
palm\_finetuned & 0.98 & 0.97 & 0.91 & 0.66 & 0.73 & 0.56 \\
\bottomrule
\end{tabular}
\caption{The percent of ``Yes'' responses for the set of articles that have summaries generated by all systems, broken down by model and question.}\label{tab:test2_yes_rate}
\end{table*}

\section{\seahorse example summaries and scores}\label{app:seahorse_examples}

\begin{figure*}
\centering
\includegraphics[width=\textwidth]{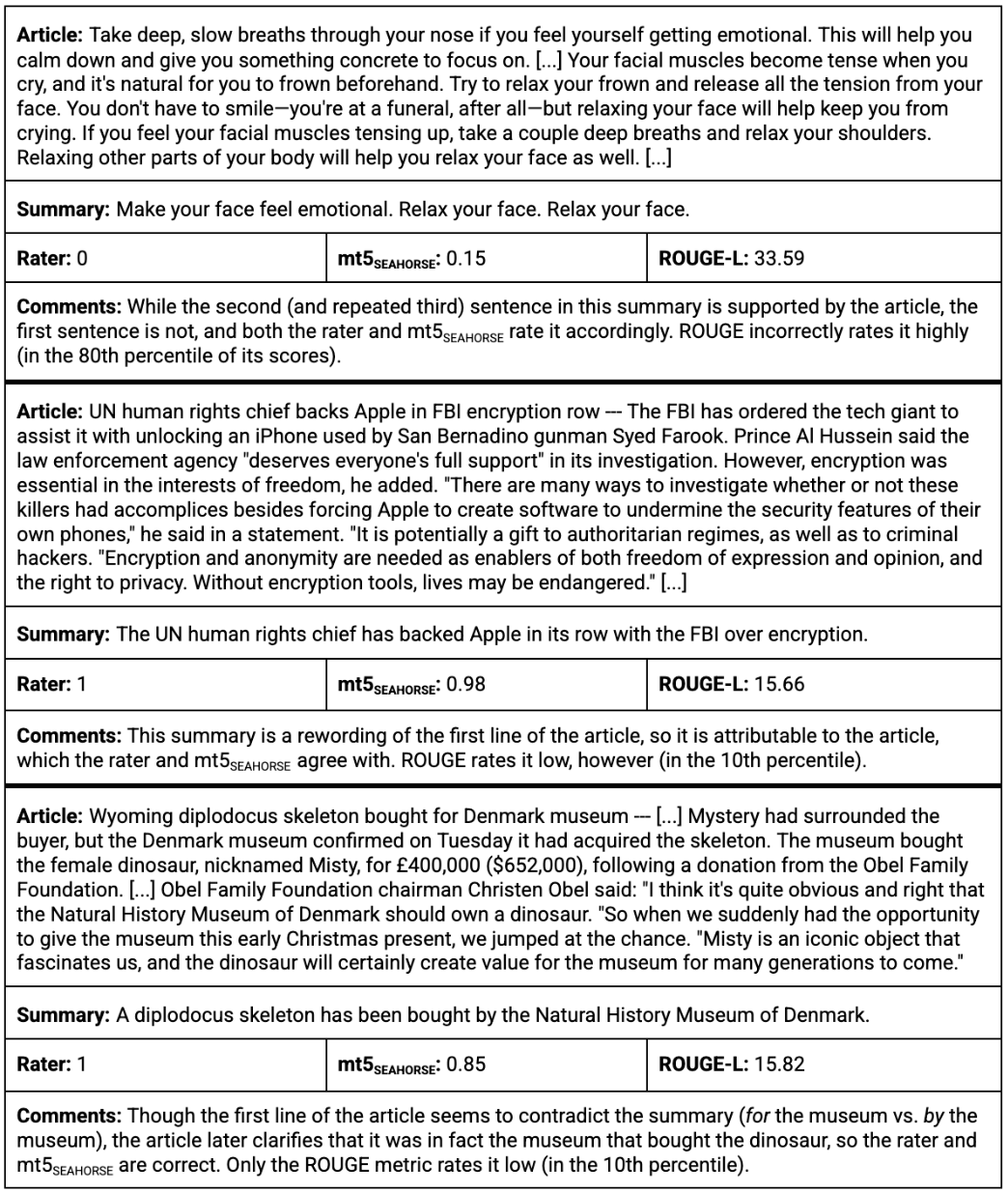}
\caption{Example summaries and ratings from the human raters, \seahorsemodel, and ROUGE-L for attribution (Q4).}
\label{fig:seahorse_examples}
\end{figure*}

\autoref{fig:seahorse_examples} shows 3 summaries from the \seahorse dataset, along with ratings for the attribution (Q4) dimension from the human raters, \seahorsemodel, and ROUGE-L.

\section{Comparison between mT5\_large and mT5\_xxl}\label{app:mt5_large}
\autoref{tab:mt5_large_results} compares the results of two versions of mT5 finetuned on \seahorse data, mT5\_large and mT5\_xxl, on the \seahorse and mFACE test sets.
Scores are generally close between the two models, but mT5\_xxl outperforms the large metric in all cases except one.
\begin{table*}[t]
\centering
\small
\begin{tabular}{l|l|ll|ll|ll|ll|ll|ll}
\toprule
& \multicolumn{1}{c|}{}     & \multicolumn{2}{c|}{Q1}                                       & \multicolumn{2}{c|}{Q2}                                       & \multicolumn{2}{c|}{Q3}                                       & \multicolumn{2}{c|}{Q4}                                       & \multicolumn{2}{c|}{Q5}                                       & \multicolumn{2}{c}{Q6}                                       \\ \hline
Dataset & Metric                    & \multicolumn{1}{c|}{$\rho$}        & \multicolumn{1}{c|}{roc} & \multicolumn{1}{c|}{$\rho$}        & \multicolumn{1}{c|}{roc} & \multicolumn{1}{c|}{$\rho$}        & \multicolumn{1}{c|}{roc} & \multicolumn{1}{c|}{$\rho$}        & \multicolumn{1}{c|}{roc} & \multicolumn{1}{c|}{$\rho$}        & \multicolumn{1}{c|}{roc} & \multicolumn{1}{c|}{$\rho$}        & \multicolumn{1}{c}{roc} \\ \hline

\seahorse & $\textrm{mt5\_L}$ & \multicolumn{1}{l|}{0.44} & 0.88            & \multicolumn{1}{l|}{0.74} & 0.97            & \multicolumn{1}{l|}{0.37} & 0.81           & \multicolumn{1}{l|}{0.55} & 0.82            & \multicolumn{1}{l|}{0.46} & 0.78            & \multicolumn{1}{l|}{0.45} & 0.77  \\
& $\textrm{mt5\_XXL}$ & \multicolumn{1}{l|}{0.52} & 0.90            & \multicolumn{1}{l|}{0.86} & 0.98            & \multicolumn{1}{l|}{0.45} & 0.84            & \multicolumn{1}{l|}{0.59} & 0.85            & \multicolumn{1}{l|}{0.50} & 0.80            & \multicolumn{1}{l|}{0.52} & 0.81           \\ \hline
mFACE - & $\textrm{mt5\_L}$ & \multicolumn{1}{l|}{0.14} & 0.77            & \multicolumn{1}{l|}{-} & -           & \multicolumn{1}{l|}{-} &-           & \multicolumn{1}{l|}{0.48} & 0.78            & \multicolumn{1}{l|}{-} & -            & \multicolumn{1}{l|}{0.32} & 0.70  \\
5 langs & $\textrm{mt5\_XXL}$ & \multicolumn{1}{l|}{0.09} & 0.73            & \multicolumn{1}{l|}{-} & -            & \multicolumn{1}{l|}{-} & -            & \multicolumn{1}{l|}{0.50} & 0.79            & \multicolumn{1}{l|}{-} & -            & \multicolumn{1}{l|}{0.50} & 0.81           \\ \hline
mFACE - & $\textrm{mt5\_L}$ & \multicolumn{1}{l|}{0.13} & 0.68            & \multicolumn{1}{l|}{-} & -            & \multicolumn{1}{l|}{-} & -          & \multicolumn{1}{l|}{0.46} & 0.77            & \multicolumn{1}{l|}{-} & -            & \multicolumn{1}{l|}{0.36} & 0.71  \\
all langs & $\textrm{mt5\_XXL}$ & \multicolumn{1}{l|}{0.15} & 0.70            & \multicolumn{1}{l|}{-} & -            & \multicolumn{1}{l|}{-} & -            & \multicolumn{1}{l|}{0.52} &0.81            & \multicolumn{1}{l|}{-} & -            & \multicolumn{1}{l|}{0.40} & 0.74           \\ \hline
\end{tabular}
\caption{Metrics' ability to predict \dataset and mFACE ratings, measured with Pearson's coefficient ($\rho$) and the area under the ROC curve (roc).
Q1 maps to ``Quality'' in the mFACE dataset, Q4 to ``Attribution,'' and Q6 to ``Informativeness.''
$\textrm{mt5\_L}$ is a \seahorse-finetuned version of mT5\_large; $\textrm{mt5\_XXL}$ is a \seahorse-finetuned version of mT5\_xxl.}\label{tab:mt5_large_results}
\end{table*}

\end{document}